%% file: main.tex
\documentclass[dvipsnames]{article} 
\usepackage{iclr2016_conference,times}
\usepackage{color}
\usepackage{hyperref}
\usepackage{url}
\usepackage{natbib}
\usepackage{graphicx}
\usepackage{amsmath}
\usepackage{amssymb}
\usepackage{caption}
\usepackage{capt-of}
\usepackage{xr} 
\usepackage{subcaption}
\usepackage{tabularx}
\graphicspath{{figures/}{../figures/}}
\usepackage{ownstyles}

\newcommand{\layer}[1]{\ensuremath{\mathsf{#1}\xspace}}

\newcommand{\net}[1]{\ensuremath{\mathsf{Net#1}\xspace}}

\newif\ifcomments

\commentsfalse

\ifcomments
\newcommand{\comments}[1]{#1}

\else
\newcommand{\comments}[1]{\ignorespaces}

\fi

\iclrfinalcopy

\title{Convergent Learning: Do different neural networks learn the same representations?}

\input{authors}

\begin{document}
\maketitle

\begin{abstract}
  Recent successes in training large, deep neural networks have prompted active investigation into the representations learned on their intermediate layers. Such research is difficult because it requires making sense of non-linear computations performed by millions of learned parameters, but
  valuable because it increases our ability to understand current models and training algorithms and thus create improved versions of them.
In this paper we investigate the extent to which neural networks exhibit
what we call {\em convergent learning}, which is when the representations learned by multiple nets converge to a set of features which are either individually similar between networks or where subsets of features span similar low-dimensional spaces.
We propose a specific method of probing representations: training multiple networks and then
comparing and contrasting their individual, learned representations at the level of neurons or groups of neurons.
We begin research into this question by introducing three techniques to approximately align different neural networks on a feature or subspace level: a bipartite matching approach that makes one-to-one assignments between neurons, a sparse prediction and clustering approach that finds one-to-many mappings, and a spectral clustering approach that finds many-to-many mappings. 
This initial investigation reveals a few interesting, previously unknown properties of neural networks, and we argue that future research into the question of convergent learning will yield many more. The insights described here include (1) that some features are learned reliably in multiple networks, yet other features are not consistently learned; (2) that units learn to span low-dimensional subspaces and, while these subspaces are common to multiple networks, the specific basis vectors learned are not; (3) that the representation codes show evidence of being a mix between a local (single unit) code and slightly, but not fully, distributed codes across multiple units; (4) that the average activation values of neurons vary considerably within a network, yet the mean activation values across different networks converge to an almost identical distribution. \footnote{A preliminary version of this work was presented at the NIPS 2015 Feature Extraction workshop.} 
\end{abstract}

\section{Introduction}

Many recent studies have focused on understanding deep neural networks from both a theoretical perspective
\citep{arora2013provable,neyshabur2013sparse,montavon2011kernel,paul2014does,goodfellow-2014-arXiv-explaining-and-harnessing-adversarial}
and from an empirical perspective
\citep{erhan2009difficulty,eigen2013understanding,  szegedy2013intriguing-properties-of-neural, simonyan2013deep-inside-convolutional, zeiler2014visualizing, nguyen2014deep, yosinski-2014-NIPS-how-transferable-are-features-in-deep, mahendran2014understanding, yosinski-2015-ICML-DL-understanding-neural-networks, zhou-2014-arXiv-object-detectors-emerge}.
In this paper we continue this trajectory toward attaining a deeper understanding of neural net training
by proposing a new approach.
We begin by noting that modern deep neural networks (DNNs) exhibit an interesting phenomenon:
networks trained starting at different random initializations frequently converge to solutions with similar performance (see \cite{dauphin2014identifying-and-attacking-the-saddle} and  \secref{setup} below).
 Such similar performance by different networks raises the question of to what extent the learned internal representations differ:
Do the networks learn radically different sets of features that happen to perform similarly, or do they exhibit \emph{convergent learning}, meaning that their learned feature representations are largely the same?
This paper makes a first attempt at asking and answering these questions. Any improved understanding of what neural networks learn should improve our ability to design better architectures, learning algorithms, and hyperparameters, ultimately enabling more capable models. For instance, distributed data-parallel neural network training is more complicated than distributed data-parallel training of convex models because periodic direct averaging of model parameters is not an effective strategy: perhaps solving a neuron correspondence problem before averaging would mitigate the need for constant synchronization. As another example, if networks converge to diverse solutions, then perhaps additional performance improvements are possible via training multiple models and then using model compilation techniques to realize the resulting ensemble in a single model.  

In this paper, we investigate the similarities and differences between the representations learned by
neural networks with the same architecture trained from different random initializations.
We employ an architecture derived from AlexNet
\citep{krizhevsky2012imagenet-classification-with-deep}
and train multiple networks on the ImageNet dataset
\citep{deng2009imagenet:-a-large-scale-hierarchical}
(details in \secref{setup}). We then compare the representations learned
across different networks. We demonstrate the effectiveness of this method by both visually and quantitatively showing that the features learned by some neuron clusters in one network can be quite similar to those learned by neuron clusters in an independently trained neural network.
Our specific contributions are asking and shedding light on the following questions:

\begin{enumerate}

\item By defining a measure of similarity between units\footnote{Note that we use the words ``filters'', ``channels'', ``neurons'', and ``units'' interchangeably to mean channels for a convolutional layer or individual units in a fully connected layer.}  in different neural networks,
  can we come up with a permutation for the units of one network to bring it into a one-to-one alignment
  with the units of another network trained on the same task? Is this matching or alignment close, because features learned by one network are learned nearly identically somewhere on the same layer of the second network, or is the approach ill-fated, because the representations of each network are unique?
  (Answer: a core representation is shared, but some rare features are learned in one network but not another; see \secref{hard}).

\vspace{-0.2em}

\item Are the above one-to-one alignment results robust with respect to different measures of neuron similarity?
  (Answer: yes, under both linear correlation and estimated mutual information metrics; see \secref{hard_mi}).
\vspace{-0.2em}

\item To the extent that an accurate one-to-one neuron alignment is
  not possible, is it simply because one network's representation
  space is a rotated version\footnote{Or, more generally, a space that
    is an affine transformation of the first network's representation
    space.} of another's? If so, can we find and characterize these
  rotations?
  (Answers: by learning a sparse weight LASSO model to predict one
  representation from only a few units of the other, we can see that 
  the transform from
  one space to the other can be possibly decoupled into transforms between
  small subspaces; see \secref{sparse}).
  \jby{Yixuan: I don't think we can make such a strong statement, and I propose changing back to the version commenting on only the first two layers}
  \later{Make and show these plots, then add comment on how the size varies by layer}

\item Can we further cluster groups of neurons from one network with a
  similar group from another network? (Answer: yes. To approximately
  match clusters, we adopt a spectral clustering algorithm
  that enables many-to-many mappings to be found between networks. See
  \secref{spectral}).

\vspace{-0.2em}

\item For two neurons detecting similar patterns, are the activation statistics similar as well?
 (Answer: mostly, but with some differences; see \secref{means}).

\end{enumerate}




\vspace*{-1em}
\section{Experimental Setup}
\seclabel{setup}
\vspace*{-.5em}

All networks in this study follow the basic architecture laid out by \cite{krizhevsky2012imagenet-classification-with-deep}, with parameters learned in five convolutional layers (\layer{conv1} -- \layer{conv5}) followed by three fully connected layers (\layer{fc6} -- \layer{fc8}). The structure is modified slightly in two ways. First, \cite{krizhevsky2012imagenet-classification-with-deep} employed limited connectivity between certain pairs of layers to enable splitting the model across two GPUs.\footnote{In \cite{krizhevsky2012imagenet-classification-with-deep} the \layer{conv2}, \layer{conv4}, and \layer{conv5} layers were only connected to half of the preceding layer's channels.}
Here we remove this artificial group structure and allow all channels on each layer to connect to all channels on the preceding layer, as we wish to study only the group structure, if any, that arises naturally, not that which is created by architectural choices. Second, we place the local response normalization layers after the pooling layers following the defaults released with the Caffe framework, which does not significantly impact performance \citep{jia2014caffe:-convolutional-architecture}.
Networks are trained using Caffe on the ImageNet Large Scale Visual Recognition Challenge (ILSVRC) 2012 dataset \citep{deng2009imagenet:-a-large-scale-hierarchical}. Further details and the complete code necessary to reproduce these experiments is available at \url{https://github.com/yixuanli/convergent_learning}.


We trained four networks in the above manner using four different random initializations. We refer to these as \net{1}, \net{2}, \net{3}, and \net{4}. The four networks perform very similarly on the validation set, achieving top-1 accuracies of
58.65\%, 58.73\%, 58.79\%, and 58.84\%,                      
which are similar to the top-1 performance of 59.3\% reported in the original study \citep{krizhevsky2012imagenet-classification-with-deep}.


We then aggregate certain statistics of the activations within the networks. Given a network \net{\emph{n}} trained in this manner,
the scalar random variable $X^{(n)}_{l,i}$ denotes the series of activation values produced over the entire ILSVRC validation dataset
by unit
$i$ on layer $l\in\{\layer{conv1},\layer{conv2},\layer{conv3},\layer{conv4},\layer{conv5},\layer{fc6},\layer{fc7}\}$.\footnote{
  For the fully connected layers, the random variable $X^{(n)}_{l,i}$ has one specific value for each input image; for the convolutional layers, the value of $X^{(n)}_{l,i}$ takes on different values at each spatial position. In other words, to sample an $X^{(n)}_{l,i}$ for an FC layer, we pick a random image from the validation set; to sample $X^{(n)}_{l,i}$ for a conv layer, we sample a random image and a random position within the conv layer.} 
 We collect the following statistics by aggregating over the validation set (and in the case of convolutional layers also over spatial positions):
\newcommand{\eqlabelstr}[1]{\textrm{#1}}
\begin{eqnarray*}
\eqlabelstr{Mean:}\hspace{2em}  \mu^{(n)}_{l,i} &=& \mathbb{E}[X^{(n)}_{l,i}] \\ 
\eqlabelstr{Standard deviation:}\hspace{2em}  \sigma^{(n)}_{l,i} &=&\sqrt (\mathbb{E}[(X^{(n)}_{l,i} - \mu^{(n)}_{l,i})^2]) \\
\eqlabelstr{Within-net correlation:}\hspace{2em}  c^{(n)}_{l,i,j} &=& \mathbb{E}[(X^{(n)}_{l,i}-\mu^{(n)}_{l,i})(X^{(n)}_{l,j}-\mu^{(n)}_{l,j})] / \sigma^{(n)}_{l,i} \sigma^{(n)}_{l,j} \\
\eqlabelstr{Between-net correlation:}\hspace{1.3em}  c^{(n,m)}_{l,i,j} &=& \mathbb{E}[(X^{(n)}_{l,i}-\mu^{(n)}_{l,i})(X^{(m)}_{l,j}-\mu^{(m)}_{l,j})] / \sigma^{(n)}_{l,i} \sigma^{(m)}_{l,j}
\end{eqnarray*}

\vspace*{-.5em}
Intuitively, we compute the mean and standard deviation of the activation of each unit in the network over the validation set. For convolutional layers, we compute the mean and standard deviation of each channel. The mean and standard deviation for a given network and layer is a vector with length equal to the number of channels (for convolutional layers) or units (for fully connected layers).\footnote{
  For reference, the number of channels for \layer{conv1} to \layer{fc8} is given by: $\mathcal{S}=\{96, 256, 384, 384, 256, 4096, 4096, 1000\}$. The corresponding size of the correlation matrix in each layer is: $\{s^2~|~\forall s \in \mathcal{S}\}$. Furthermore, the spatial extents of each channel in each convolutional layer is given by: $\{\layer{conv1}:55\times55, \layer{conv2}:27\times27, \layer{conv3}: 13\times13,\layer{conv4}: 13\times13,\layer{conv5}: 13\times13\}$}
The within-net correlation values for each layer can be considered as a symmetric square matrix with side length equal to the number of units in that layer (e.g. a $96\times96$ matrix for \layer{conv1} as in \figref{cor_and_xcor_conv1}a,b). For a pair of networks, the between-net correlation values also form a square matrix, which in this case is not symmetric (\figref{cor_and_xcor_conv1}c,d).

\begin{figure}[t]
\begin{center}
  \includegraphics[width=1.0\linewidth]{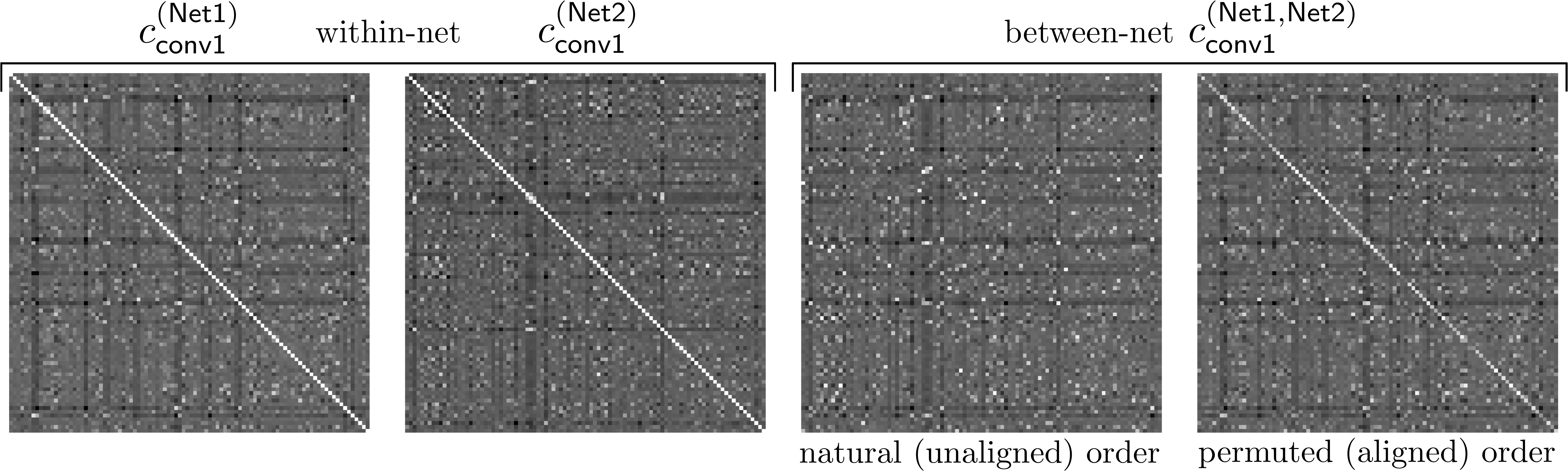}\\
  (a)\hspace{.225\linewidth}(b)\hspace{.225\linewidth}(c)\hspace{.225\linewidth}(d)
  \caption{Correlation matrices for the \layer{conv1} layer, displayed as images with minimum value at black and maximum at white. \textbf{(a,b)} Within-net correlation matrices for \net{1} and \net{2}, respectively. \textbf{(c)} Between-net correlation for \net{1} vs. \net{2}. \textbf{(d)} Between-net correlation for \net{1} vs. a version of \net{2} that has been permuted to approximate \net{1}'s feature order. The partially white diagonal of this final matrix shows the extent to which the alignment is successful; see \figref{match_vs_max_conv1} for a plot of the values along this diagonal and further discussion.
  }
  \figlabel{cor_and_xcor_conv1}
\end{center}
\vspace{-1em}
\end{figure}

We use these correlation values as a way of measuring how related the activations of one unit are to another unit, either within the network or between networks. We use correlation to measure similarity because it is independent of the scale of the activations of units.
Within-net correlation quantifies the similarity between two neurons in the same network; whereas the between-net correlation matrix quantifies the similarity of two neurons from different neural networks.
Note that the units compared are always on the same layer on the network; we do not compare units between different layers.
In the Supplementary Information (see \figref{activation_correlation}),
we plot the activation values for several example high correlation and low correlation pairs of units from \layer{conv1} and \layer{conv2} layers; the simplicity of the distribution of values suggests that the correlation measurement is an adequate indicator of the similarity between two neurons. 
To further confirm this suspicion, we also tested with a full estimate of the mutual information between units and found it to yield similar results to correlation (see \secref{hard_mi}). 

\section{Is There a One-to-One Alignment Between Features Learned by Different Neural Networks?}

\seclabel{hard}



We would like to investigate the similarities and differences between multiple training runs of same network
architecture. Due to symmetries in the architecture and weight initialization procedures, for any given parameter
vector that is found, one could create many equivalent solutions simply by permuting the unit orders within a layer (and permuting
the outgoing weights accordingly). 
Thus, as a first step toward analyzing the similarities and differences
between different networks, we ask the following question: if we allow ourselves to permute the units of one network,
to what extent can we bring it into alignment with another?
To do so requires finding equivalent or nearly-equivalent units across networks, and for this task we adopt the magnitude independent measures of correlation and mutual information. We primarily give results with the simpler, computationally faster correlation measure (\secref{hard_correlation}), but then confirm the mutual information measure provides qualitatively similar results (\secref{hard_mi}).

\subsection{Alignment via Correlation}
\seclabel{hard_correlation}

As discussed in \secref{setup}, we compute within-net and between-net unit correlations.
\figref{cor_and_xcor_conv1} shows the within-net correlation values computed between units on a network and other units on the same network (panels a,b) as well as the between-net correlations between two different networks (panel~c).
We find matching units between a pair of networks --- here \net{1} and \net{2} --- in two ways. In the first approach, for each unit in \net{1}, we find the unit in \net{2} with maximum correlation to it, which is the max along each row of \figref{cor_and_xcor_conv1}c.
This type of assignment is known as a bipartite \emph{semi-matching} in graph theory \citep{lawler-1976-combinatorial-optimization:-networks}, and we adopt the same nomenclature here. This procedure can result in multiple units of \net{1} being paired with the same unit in \net{2}. 
 \figref{match_ims_top_bot} shows the eight highest correlation matched features and eight lowest correlation matched features 
using the semi-matching approach (corresponding to the leftmost eight and rightmost eight points in \figref{match_vs_max_conv1}). To visualize the functionality each unit, we plot the image patch from the validation set that causes the highest activation for that unit. 
For all the layers shown, the most correlated filters (on the left) reveal that there are nearly perfect counterparts in each network, whereas the low-correlation filters (on the right) reveal that there are many features learned by one network that are unique and thus have no corollary in the other network. 

\begin{figure}[t]
  \vspace{-0em}
\begin{center}
  \includegraphics[width=1\linewidth]{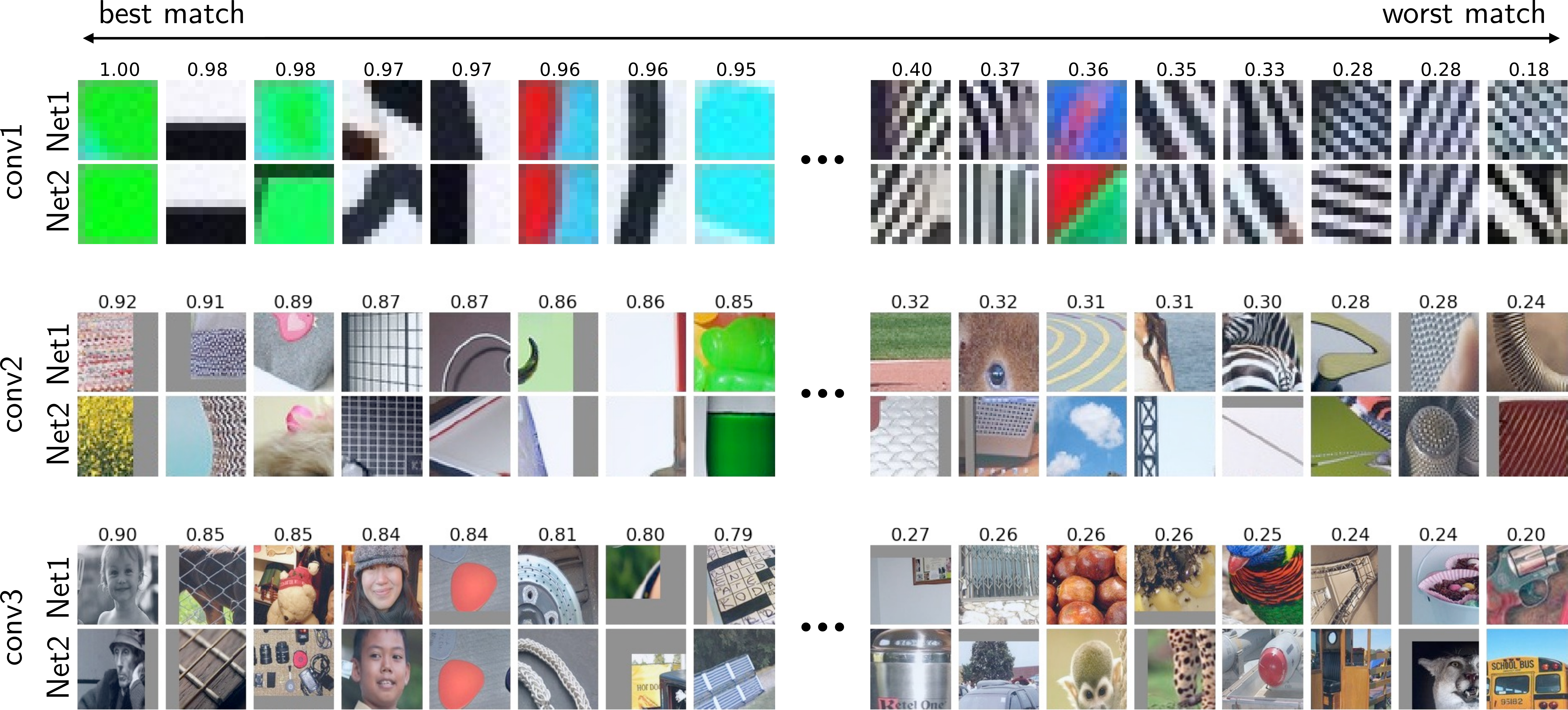}
  \caption{ With assignments chosen by semi-matching, the eight best (highest correlation, left) and eight worst (lowest correlation, right) matched features between \net{1} and \net{2} for the \layer{conv1} -- \layer{conv3} layers. For all layers visualized, (1) the most correlated filters are near perfect matches, showing that many similar features are learned by independently trained neural networks, and (2) the least correlated features show that many features are learned by one network and are not learned by the other network, at least not by a single neuron in the other network. The results for the \layer{conv4} and \layer{conv5} layers can be found in the Supplementary Material (see \figref{match_ims_top_bot_conv4_conv5}).  
  }
  \figlabel{match_ims_top_bot}
\end{center}
\end{figure}

An alternative approach is to find the one-to-one assignment between units in \net{1} and \net{2} without replacement, such that every unit in each network is paired with a unique
unit in the other network. This more common approach is known as bipartite \emph{matching}.\footnote{Note that the \emph{semi-matching} is ``row-wise greedy'' and will always have equal or better sum of correlation than the \emph{matching}, which maximizes the same objective but must also satisfy global constraints.} A matching that maximizes the sum
of the chosen correlation values may be found efficiently
via the Hopcroft-Karp algorithm \citep{hopcroft-1973-an-n5/2-algorithm-for-maximum} after turning the between-net correlation matrix into a weighted bipartite graph.
\figref{cor_and_xcor_conv1}c shows an example between-net correlation matrix;
the max weighted matching can be thought of as a path through the matrix such that each row and each column are selected exactly once, and the sum of elements along the path is maximized.
Once such a path is found, we can permute the units of \net{2} to bring it into the best possible alignment with \net{1}, so that the first channel of \net{2} approximately matches (has high correlation with) the first channel of \net{1}, the second channels of each also approximately match, and so on.
The correlation matrix of \net{1} with the permuted version of \net{2} is shown in \figref{cor_and_xcor_conv1}d. Whereas the diagonal of the self correlation matrices are exactly one, the diagonal of the permuted between-net correlation matrix contains values that are generally less than one. Note that the diagonal of the permuted between-net correlation matrix is bright (close to white) in many places, which shows that for many units in \net{1} it is possible to find a unique, highly correlated unit in \net{2}.

\figref{match_vs_max_conv1} shows a comparison of assignments produced by the semi-matching and matching methods for the \layer{conv1} layer (\figref{match_vs_max_all} shows results for other layers). 
%
Insights into the differing representations learned can be gained from both assignment methods.
The first conclusion is that for most units, particularly those with the higher semi-matching and matching correlations (\figref{match_vs_max_conv1}, left), the semi-matching and  matching assignments coincide, revealing that for many units a one-to-one assignment is possible.
\jmc{That right? The fact that semi and matching are the same days nothing about one-to-one (vs. many-to-one), but just about whether replacement helps, no?}
\jby{address later}
%
Both methods reveal that the average correlation for one-to-one alignments varies from layer to layer (\figref{average_correlation}), with the highest matches in the \layer{conv1} and \layer{conv5} layers, but worse matches in between. This pattern implies
that the path from 
a relatively matchable \layer{conv1} representation to \layer{conv5} representation passes
through an intermediate middle region where matching is more difficult, suggesting that what is learned by different networks on \layer{conv1} and \layer{conv2} is more convergent than \layer{conv3}, \layer{conv4} and \layer{conv5}. This result may be related to previously
observed greater complexity in the intermediate layers as measured through the lens of optimization difficulty \citep{yosinski-2014-NIPS-how-transferable-are-features-in-deep}.\jmc{This previous sentence needs to be improved. I get the impression the author knows what he means, but didn't have the energy to fully write out exactly what this insight is. I don't fully get what you mean, and I know both of these papers very well, so a naive reader will have no idea precisely what we mean.}

Next, we can see that where the semi-matching and matching differ, the matching is often much worse.\jmc{this sentence is tautological, if the 2nd matching is defined as when the differ. or do you mean something else by the 2nd matching, such as "when matching scores are different, it also tends to be the case that the discrepancy between the semi-matching methods and the matching methods is higher?} 
One hypothesis for why this occurs
is that the two networks learn different numbers of units to span certain subspaces. For example, \net{1} might learn a representation that uses six filters to span a subspace of human faces, but \net{2} learns to span the same subspace with five filters. With unique matching, five out of the six filters from \net{1} may be matched to their nearest counterpart in \net{2}, but the sixth \net{1} unit will be left without a counterpart and will end up paired with an almost unrelated filter.\jmc{I think that's a non-sequitur with the first/topic sentence of this paragraph, no? Why does this fact relate to that claim? In general, this paragraph needs to be improved for clarity.}

Finally, with reference to \figref{match_vs_max_conv1} 
 (but similarly observable in \figref{match_vs_max_all} for other layers),
another salient observation is that the correlation of the semi-matching falls significantly from the best-matched unit (correlations near 1) to the lowest-matched (correlations near 0.3).
This indicates that some filters in \net{1} can be paired up with filters in \net{2} with high correlation, but other filters in \net{1} and \net{2} are network-specific and have no high-correlation pairing in the alternate network, implying that those filters are rare and not always learned.
This holds across the \layer{conv1} -- \layer{conv5} layers. 

\begin{table}[t]
    \begin{minipage}[b]{.53\textwidth}
        \includegraphics[width=1\textwidth]{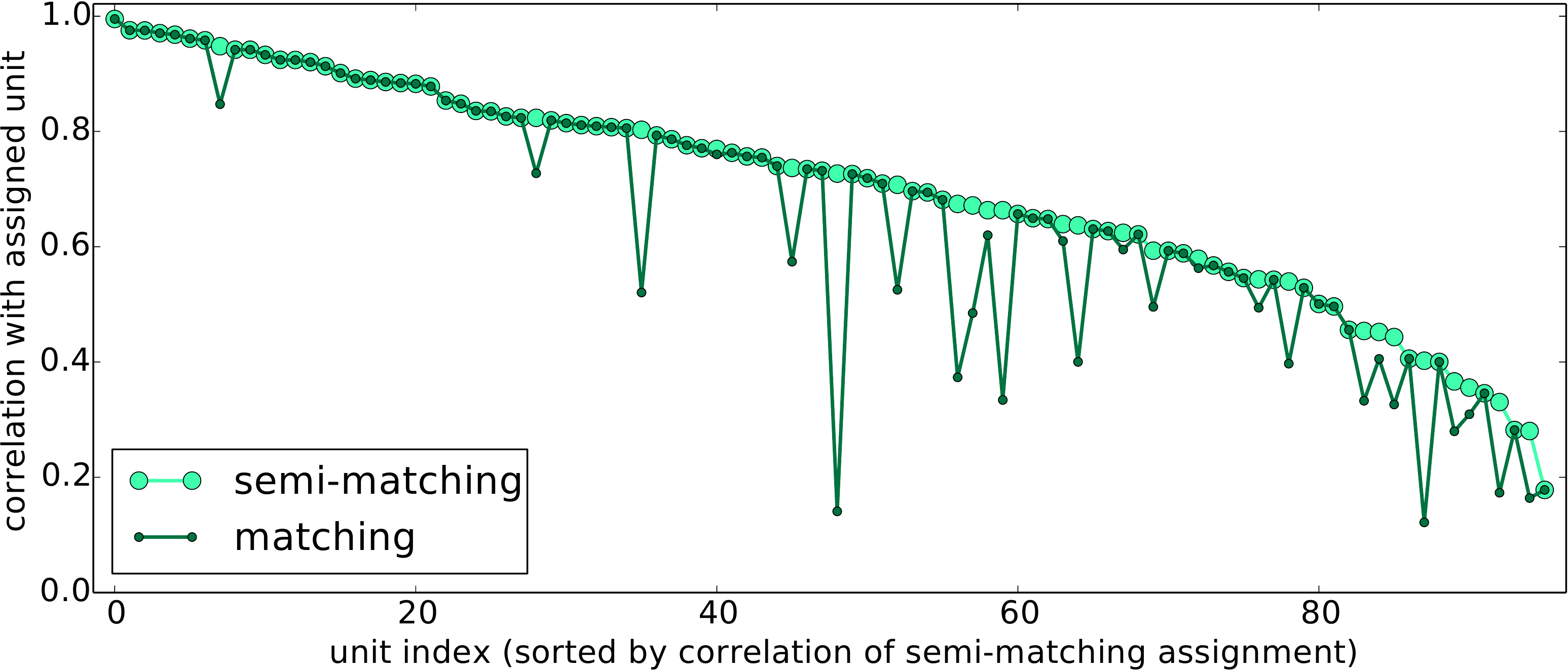}
        \captionof{figure}{Correlations between paired \layer{conv1} units in  \net{1} and \net{2}. Pairings are made via semi-matching (light green), which allows the same unit in \net{2} to be matched with multiple units in \net{1}, or matching (dark green), which forces a unique \net{2} neuron to be paired with each \net{1} neuron. Units are sorted by their semi-matching values. See text for discussion.}
        \figlabel{match_vs_max_conv1}
    \end{minipage}
    \hfill
    \begin{minipage}[b]{.41\textwidth}
        \includegraphics[width=1\textwidth]{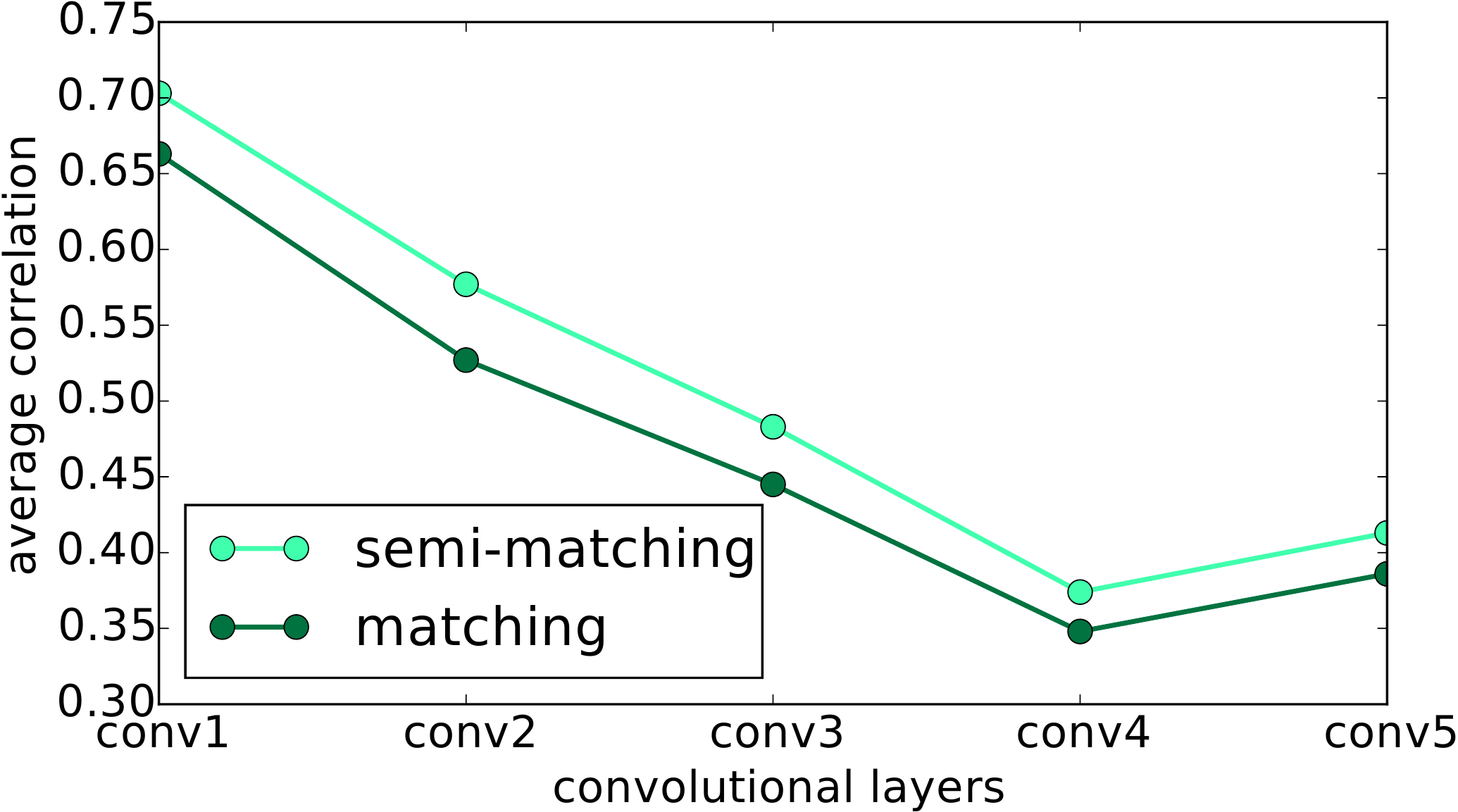}
        \captionof{figure}{Average correlations between paired \layer{conv1} units in  \net{1} and \net{2}. Both semi-matching (light green) and matching (dark green) methods suggest that features learned in different networks are most convergent on \layer{conv1} and least convergent on \layer{conv4}. 
        }
        \figlabel{average_correlation}
    \end{minipage}
\end{table}

\vskip -0.05in
\subsection{Alignment via Mutual Information}
\seclabel{hard_mi}
\vskip -0.03in
Because correlation is a relatively simple mathematical metric that may miss some forms of statistical dependence,
we also performed one-to-one alignments of neurons by measuring the {\em mutual information} between them.
Mutual information measures how much knowledge one gains about one variable by knowing the value of another.
Formally, the mutual information of the two random variables $X^{(n)}_{l,i}$and $ X^{(m)}_{l,j}$ representing the activation of the $i$-th neuron in \net{\emph{n}} and the $j$-th neuron in \net{\emph{m}}, is defined as:
$$
I\Big( X^{(n)}_{l,i};X^{(m)}_{l,j}\Big) = \sum_{a\in X^{(n)}_{l,i}} \sum_{b\in X^{(m)}_{l,j}} p(a,b) \log\Big(\frac{p(a,b)}{p(a)p(b)}\Big),
$$where $p(a,b)$ is the joint probability distribution of $X^{(n)}_{l,i}$ and $X^{(m)}_{l,j}$, and $p(a)$ and $p(b)$ are their marginal probability distributions, respectively. 
The within-net mutual information matrix and between-net mutual information matrix have the same shapes as their equivalent correlation matrices.

We apply the same matching technique described in \secref{hard_correlation} to the between-net mutual information matrix,%
\footnote{The mutual information between each pair of neurons is estimated using 1D and 2D histograms of paired activation values over 60,000 random activation samples. We discretize the activation value distribution into percentile bins along each dimension, each of which captures 5\% of the marginal distribution mass. We also add a special bin with range $(-\inf, 10^{-6}]$ in order to capture the significant mass around 0.}
and compare the highest and lowest mutual information matches (\figref{match_mi_top_bot}) to those obtained via correlation (\figref{match_ims_top_bot}). The results are qualitatively the same. For example, seven out of eight best matched pairs in the \layer{conv1} layer stay the same. These results suggest that correlation is an adequate measurement of the similarity between two neurons, and that switching to a mutual information metric would not qualitatively change the correlation-based conclusions presented above. 

\section{Relaxing the One-to-One Constraint to Find Sparse, Few-to-One Mappings}
\seclabel{sparse}


The preceding section showed that, while some neurons have a one-to-one match in another network, for other neurons no one-to-one match exists (with correlation above
some modest threshold). For example, 17\% of \layer{conv1} neurons in \net{1} have no match in \net{2} with a correlation above $0.5$ (\figref{match_vs_max_conv1}); this number rises to 37\% for \layer{conv2}, 63\% for \layer{conv3}, and 92\% for \layer{conv4}, before dropping to 75\% for \layer{conv5} (see Figure \ref{fig:match_vs_max_all}).
\later{measure more exactly and report numbers for all layers here.}
\later{Compute the expected value of overlap given random rotations for each dimension size. Try without and with relu.}

These numbers indicate that, particularly for intermediate layers, a simple one-to-one mapping
is not a sufficient model to predict the activations of some neurons in one network given the activations of neurons in another network (even with the same architecture trained on the same task).
That result could either be because the representations are unique (i.e. not convergent), or because the best possible one-to-one mapping is insufficient to tell the complete story of how one representation is related
to another.
\todo{fix that last sentence... it's unclear. Probably Jason's fault}
\later{draw the subspace hypothesis picture and refer to it here.}
We can think of a one-to-one mapping as a model that predicts
activations in the second network by multiplying the activations of the first by a permutation matrix --- a square matrix constrained such that each row and each column contain a single one and the rest zeros.
Can we do better if we learn a model without this constraint?

We can relax this one-to-one constraint to various degrees by learning
a \emph{mapping layer} with an L1 penalty (known as a LASSO model,
\citep{tibshirani-1996-regression-shrinkage-and-selection}), where
stronger penalties will lead to sparser (more few-to-one or
one-to-one) mappings. This sparsity pressure can be varied from quite
strong (encouraging a mostly one-to-one mapping) all the way to zero,
which encourages the learned linear model to be dense.
More
specifically, to predict one layer's representation from another, we
learn a single mapping layer from one to the other (similar to the ``stitching layer'' in \cite{lenc-2015-understanding-image-representations}). In the case of
convolutional layers, this mapping layer is a convolutional layer
with $1\times1$ kernel size and number of output channels equal to the
number of input channels. The mapping layer's parameters can be considered as a square weight matrix with side length equal to the number of units in the layer; the layer learns to predict any unit in one network via a linear weighted sum of any number of units in the other. 
The model and resulting square weight
matrices are shown in \figref{stitching_architecture}. 
This layer is
then trained to minimize the sum of squared prediction errors plus
an L1 penalty, the strength of which is varied.\footnote{Both representations
  (input and target) are taken after the relu is applied. Before
  \jmc{LASSO (if we are only referring to lasso here...if both lasso and spectral clustering, we need another adjective....the reason is to differentiate the mapping layer training from the original DNN training} training, each channel is normalized to have mean zero and standard
  deviation $1/\sqrt{N}$, where $N$ is the number of dimensions of the
  representation at that layer (e.g. $N = 55\cdot55\cdot96 = 290400$
  for \layer{conv1}).  This normalization has two effects.
  First, the
  channels in a layer are all given equal importance, without which the channels
  with large activation values (see \figref{means_nets1234}) dominate
  the cost and are predicted well at the expense of less
  active channels, a solution which provides little information about
  the less active channels.
  Second, the representation at any layer
  for a single image becomes approximately unit length, making the
  initial cost about the same on all layers and allowing the same
  learning rate and SGD momentum hyperparameters to be used for all
  layers. It also makes the effect of specific L1 multipliers approximately commensurate and
  allows for rough comparison of prediction performance between layers, because the scale is constant.
  }

%

\begin{table}[t]
  \centering
  \begin{tabular}{ c | c  c  c c c c} 
    \hline
    \hline
    & \multicolumn{6}{c}{{\bf Sparse Prediction Loss (after 4,500 iterations)}} \\
    & {decay 0} & {decay $10^{-5}$} & {decay $10^{-4}$} & { decay $10^{-3}$} & {decay $10^{-2}$}  &  {decay $10^{-1}$} \\
    \hline
    \layer{conv1} & {\bf 0.170} & {\bf 0.169}  & {\bf 0.162} & {\bf 0.172} & 0.484 &  0.517\\
    \layer{conv2} & {\bf 0.372} & {\bf 0.368}  & {\bf 0.337} & {\bf 0.392} & 0.518 &  0.514\\
    \layer{conv3} & {    0.434} & 0.427        & 0.383       & {    0.462} & 0.497 &  0.496\\
    \layer{conv4} & {    0.478} & 0.470        & 0.423       & {    0.477} & 0.489 &  0.488\\
    \layer{conv5} & {    0.484} & 0.478        & 0.439       & {    0.436} & 0.478 &  0.477\\
    \hline
    \hline
  \end{tabular}
  \caption{
    Average prediction error for mapping layers with varying L1 penalties (i.e. decay terms). Larger decay parameters enforce stronger sparsity in the learned weight matrix.
    Notably, on \layer{conv1} and \layer{conv2}, the prediction errors do not rise much
    compared to the dense (decay = 0) case
    with the imposition of a sparsity penalty until after an L1 penalty weight of over $10^{-3}$ is used.
    This region of roughly constant performance despite increasing sparsity pressure is shown in bold.
    That such extreme sparsity does not hurt performance implies that each neuron in one network can be predicted by only one or a few neurons in another network. For the \layer{conv3}, \layer{conv4}, and \layer{conv5} layers, the overall error is higher, so it is difficult to draw any strong conclusions regarding those layers. The high errors could be because of the uniqueness of the learned representations, or the optimization could be learning a suboptimal mapping layer for other reasons.
    \later{Give percent of non-zero weights for each layer/L1 penalty}
    \later{Plot percent of non-zero weights for each layer/L1 penalty}
    \later{Try learning a model with one or two hidden layers to see if conv3/4/5 results improve.}
  }
  \tablabel{sparse_losses}
  \vspace{-0em}
\end{table}

Mapping layers are trained for layers \layer{conv1} -- \layer{conv5}. The average squared prediction
errors for each are shown in \tabref{sparse_losses} for a variety of L1 penalty weights (i.e. different decay values). For the \layer{conv1} and \layer{conv2} layers, the prediction errors do not rise with the imposition of a sparsity penalty until a penalty greater than $10^{-3}$. A sparsity penalty as high as $10^{-3}$ results in mapping layer models that are nearly as accurate as their dense counterparts, but that contain mostly zero weights. Additional experiments for \layer{conv1} revealed that a penalty multiplier of $10^{-2.6}$ provides a good trade-off between sparsity and accuracy, resulting in a model with sparse prediction loss 0.235, and an average of 4.7 units used to predict each unit in the target network (i.e. an average of 4.7 significantly non-zero weights). For \layer{conv2} the
$10^{-3}$ multiplier worked well, producing a model with an average of 2.5 non-zero connections per predicted unit.
The mapping layers for higher layers (\layer{conv3} -- \layer{conv5}) showed poor performance even without regularization, for reasons we do not yet fully understand,
so further results on those layers are not included here. Future investigation with different hyperparameters or different architectures (e.g. multiple hidden layers) could train more powerful predictive models for these layers.

The one-to-one results of \secref{hard} considered in combination with these results on
sparse prediction shed light on the open, long-standing debate about the extent to which learned representations are local vs. distributed~\todo{cite dump, including Bengio's monograph and neuroscientist papers}:
The units that match well one-to-one suggest
the presence of a \emph{local} code, where each of these dimensions is important enough, independent enough, or privileged enough in some other way to be relearned by different networks. 
Units that do not match well one-to-one, but are predicted well by a sparse model, suggest the presence, along those dimensions, of \emph{slightly distributed} codes. \jmc{do we want to say "clusters of slightly distributed codes"?}\jmc{do we have any ability to help the reader (or us) understand the relative fraction of neurons that fall into these two camps? And are there other camps?}

The results could have been otherwise: if all units could accurately be matched one-to-one, we would suspect a local code across the whole layer. 
On the other hand, if making predictions from one network to another required a dense affine transformation, then we would interpret the code as \emph{fully distributed}, with each unit serving only as a basis vector used to span a large dimensional subspace, whose only requirement was to have large projection onto the subspace (to be useful) and small projection onto other basis vectors (to be orthogonal). \jmc{you're going to have to unpack/spoonfeed the previous sentence much more if you want people like me to understand}
The story that actually emerges is that the first two layers use a mix of a local and a slightly distributed code.
\todo{perhaps weaken this conclusion?}

\figref{stitching_conv1_result} shows a visualization of the learned weight matrix for \layer{conv1},
along with a permuted weight matrix that aligns units from \net{2} with the \net{1} units that most predict them.
We also show two example units in \net{2} and, for each, the only three units in \net{1} that are needed to predict their activation values.

\begin{figure}[t]
\begin{center}
  \includegraphics[width=0.8\linewidth]{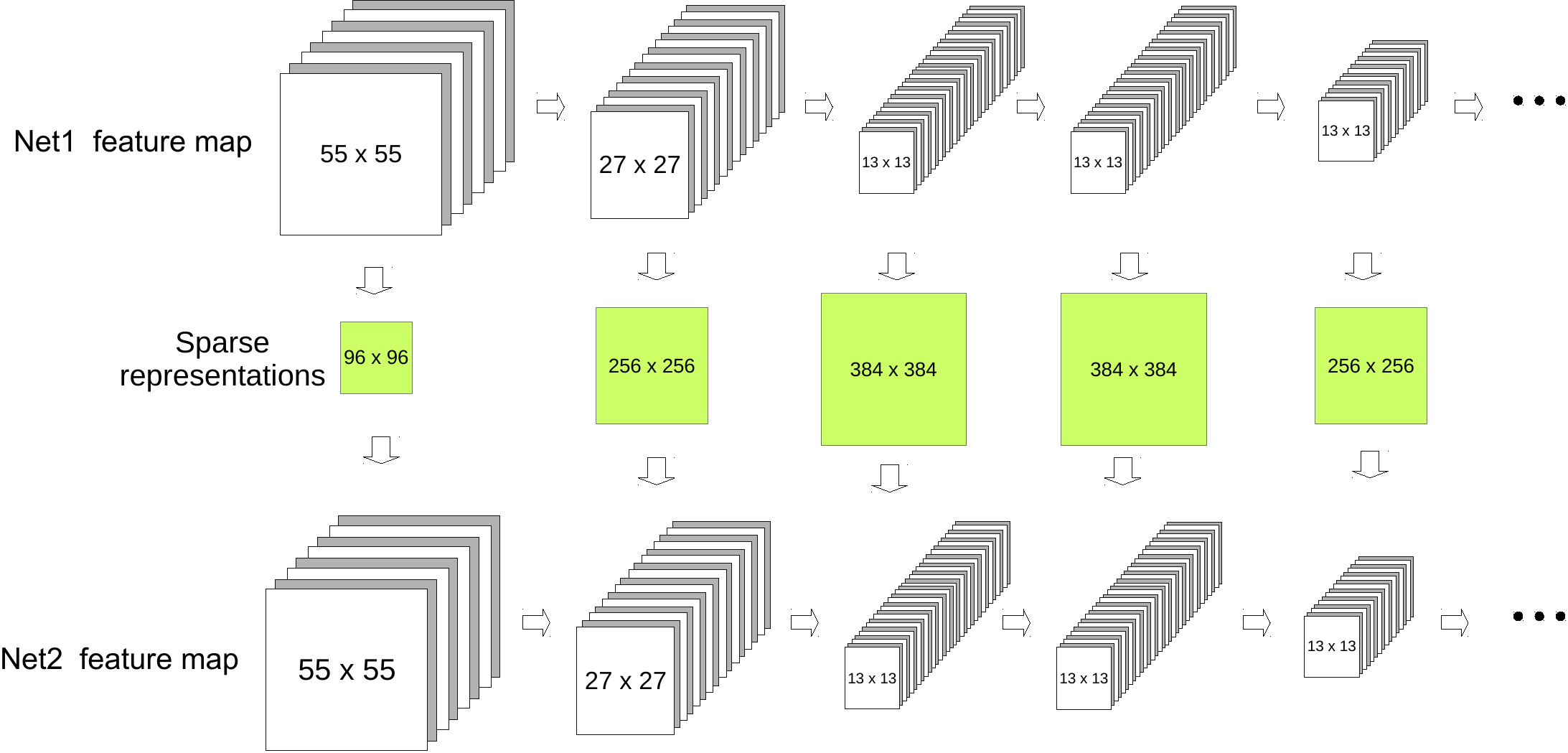}
  \caption{
    A visualization of the network-to-network sparse ``mapping layers'' (green squares). The layers are trained independently of each other and
    with an L1 weight penalty to encourage sparse weights.
}
  \figlabel{stitching_architecture}
\end{center}
\vspace{-1em}
\end{figure}

These sparse prediction results suggest that small groups of units in each network span similar subspaces, but we have not yet identified or visualized the particular subspaces that are spanned. Below we present one approach to do so by using the sparse prediction matrix directly, and
in the Supplementary \secref{spectral} we discuss a related approach using spectral clustering.

For a layer with $s$ channels we begin by creating a $2s\times2s$ block matrix $B$ by concatenating the blocks $[I,~W;~W^T,~I]$ where $I$ is the $s\times s$ identity matrix and $W$ is the learned weight matrix. 
Then
we use Hierarchical Agglomerative Clustering (HAC), as implemented in Scikit-learn \citep{pedregosa-2011-scikit-learn:-machine-learning}
to recursively cluster individual units into 
 clusters, and those clusters into larger \jmc{can we be more specific, or does the algorithm determine the size of the clusters such that we can't specify?} clusters, until all have been joined
into one cluster. The HAC algorithm as adapted to this application works in the following way:
(1) For all pairs of units, we find the biggest
off-diagonal value in B, i.e. the largest prediction weight;
(2) We pair those two units together into a cluster and consider it as a single entity for the remainder of the algorithm;
(3) We start again from step 2 using the same process (still looking for the
biggest value), but whenever we need to compare unit~$\leftrightarrow$~cluster or cluster~$\leftrightarrow$~cluster, we use the average unit~$\leftrightarrow$~unit weight over all cross-cluster pairs;\jmc{I'd love some spoon-feeding info here about whether/why this is a good/valid/acceptable idea}
(4) Eventually the process terminates when there is a single cluster.\footnote{For example,
in the \layer{conv1} layer with $96$ channels, this happens after $96\cdot2-1=191$ steps.}

The resulting clustering can be interpreted as a tree with units as leaves,
clusters as intermediate nodes, and the single final cluster as the
root node. In \figref{stitching_conv1_result_tree} we plot the $B$
matrix with rows and columns permuted together in the order leaves are
encountered when traversing the tree, and intermediate nodes are
overlaid as lines joining their subordinate units or clusters. For
clarity, we color the diagonal pixels (which all have value one
) with
green or red if the associated unit came from \net{1} or
\net{2}, respectively.

Plotted in this way, structure is readily visible: Most parents of leaf clusters (the smallest merges shown as two blue lines of length two covering a $2\times2$ region) contain one unit from \net{1} and one from \net{2}. These units can be considered most predictive of each other.\footnote{Note that the upper right corner of $B$ is $W = W_{1\rightarrow 2}$, the matrix predicting \net{1} $\rightarrow$ \net{2}, and the lower left is just the transpose $W^T_{1\rightarrow 2}$. The corners could instead be $W_{1\rightarrow 2}$ and $W_{2\rightarrow 1}$, respectively.} Slightly higher level clusters show small subspaces, comprised of multiple units from each network, where multiple units from one network are useful for predicting activations from the other network \jmc{is the clustering only based on between-net predictive power, or can it also key on within-net?}\done{We could alternatively replace the identity matrix with within-net sparsity prediction, which would be more fun. For now, yes it is solely based on the between-net.} (see the example zoomed regions on the right side of \figref{stitching_conv1_result_tree}).
\todo{how do we know that these small clusters (e.g. level two clusters, which are clusters of clusters) represent subspaces? If they were orthogonal basis vectors within a subspace, wouldn't they not be correlated?}

The HAC method employs greedy merges, which could in some cases be suboptimal. In the Supplementary \secref{spectral} we explore a related method that is less greedy, but operates on the denser correlation matrix instead.
Future work investigating or developing other methods for analyzing the structure of the sparse prediction matrices may
shed further light on the shared, learned subspaces of independently trained networks.

\todo{Jason: add complete graph drawn with graphviz to SI for one or more layers}

\begin{figure}[t]
\begin{center}
  \includegraphics[width=1\linewidth]{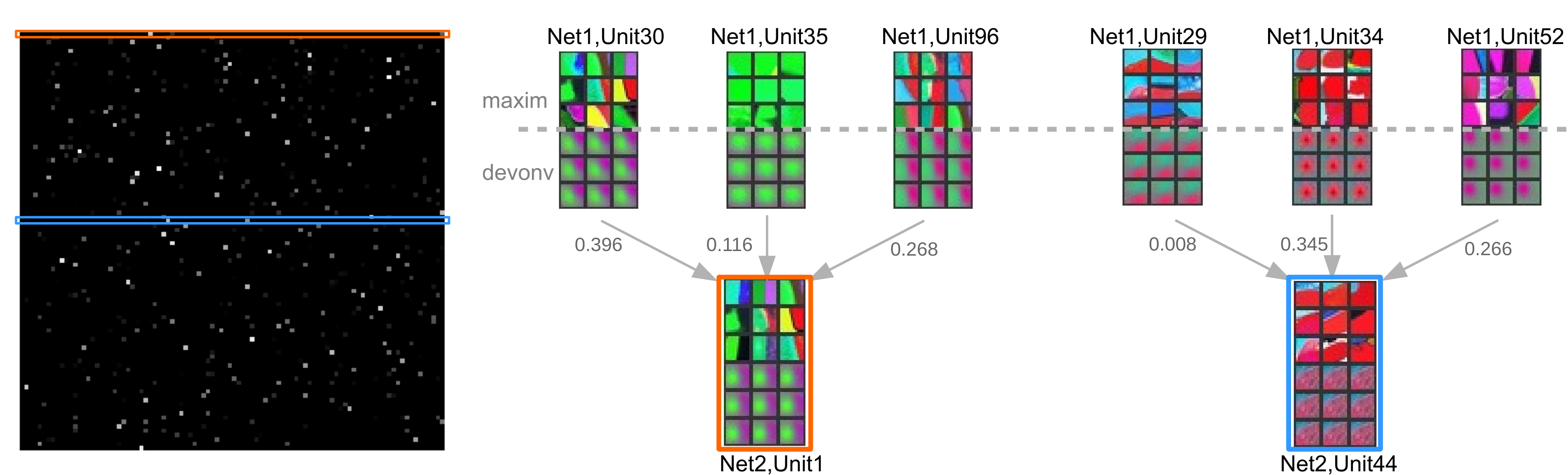}
  \caption{
    \textbf{(left)}
    The learned mapping layer from \net{1} to \net{2} for the \layer{conv1} layer.
    \textbf{(right)}
    Two example units (bottom)
    in \net{2} --- which correspond to the same colored rows in the left weight matrix --- together with, for each, the only three units  in \net{1} that are needed
    to predict their activation.
To fully visualize the functionality each unit, we plot the top 9 image patches from the validation set that causes the highest activation for that unit (``maxim''), as well as its corresponding  ``deconv'' visualization introduced by \cite{zeiler2014visualizing}. We also show the actual weight associated with each unit in \net{1} in the sparse prediction matrix.   
    \todo{would be cool/informative to add the orange and blue rectangles to the permuted matrix too}
  }
  \figlabel{stitching_conv1_result}
\end{center}
\end{figure}

\begin{figure}[th]
\begin{center}
  \includegraphics[width=1\linewidth]{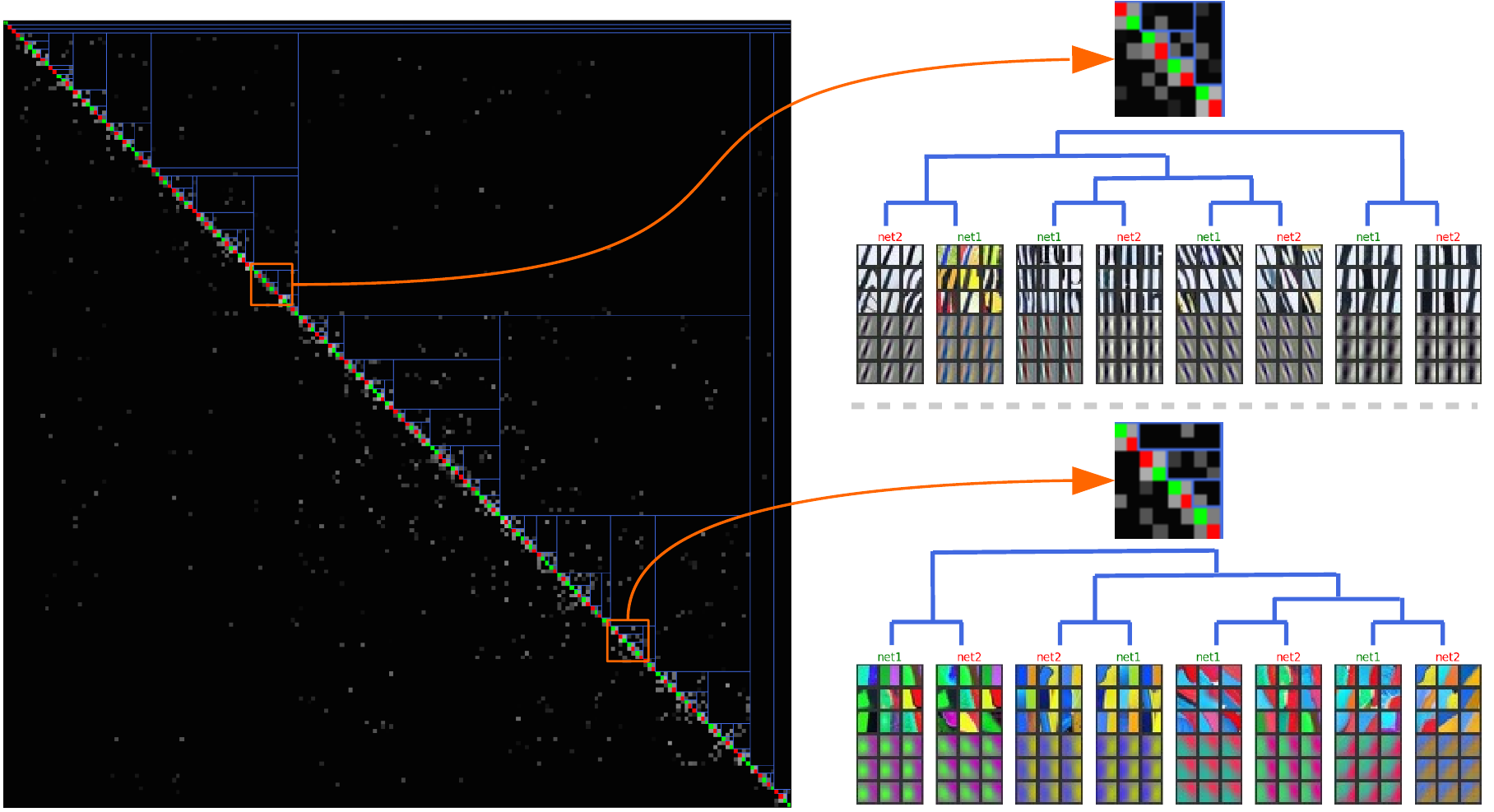}
  \caption{
    The results of the Hierarchical Agglomerative Clustering (HAC) algorithm described in \secref{sparse} on the \layer{conv1} layer.
    Left: The $B$ matrix permuted by the tree-traversal order of leaf nodes. Pixels on the diagonal are leaf nodes and represent original units of either network (green for \net{1} and red for \net{2}). The brighter the gray pixel is, the larger the weight is in the matrix. 
    See text for a complete interpretation. Right: Two zoomed in regions of the diagonal, showing two different four-dimensional subspaces spanned by four units in each network. The top 9 and bottom 9 images correspond to the maxim and deconv visualizations, respectively. 
    Best viewed digitally with zoom.
    \todo{Is this with or without the mat**.5 hack? If with: include in caption.}
    \todo{the dark blue of the tree is nearly impossible to see. How about neon blue?}
    \jmc{can we show somehow that the one-to-one matches for these pairs are not that good, but that things get better because we are using a group of them? it actually does look like the one-to-one matches here are quite good...at least for the top example}
  }
  \figlabel{stitching_conv1_result_tree}
\end{center}
\end{figure}

\vspace*{-.8em}
\section{Conclusions}
\seclabel{conclusions}
\vspace*{-.4em}

We have demonstrated a method for quantifying the feature similarity between different, independently trained deep neural networks. We show how insights may be gain by approximately aligning different neural networks on a feature or subspace level by blending three approaches: a bipartite matching that makes one-to-one assignments between neurons, a sparse prediction and clustering approach that finds one-to-many mappings, and a spectral clustering approach that finds many-to-many mappings.  
Our main findings include:
\begin{enumerate}
\item Some features are learned reliably in multiple networks, yet other features are not consistently learned.
\item Units learn to span low-dimensional subspaces and, while these subspaces are common to multiple networks, the specific basis vectors learned are not.
\item The representation codes are a mix between a local (single unit) code and slightly, but not fully, distributed codes across multiple units.
\item The average activation values of neurons vary considerably within a network, yet the mean activation values across different networks converge to an almost identical distribution.
\end{enumerate}
\vspace*{-.8em}
\section{Future Work}
\seclabel{future}
\vspace*{-.4em}
The findings in this paper open up new future research directions, for example: (1) Model compression. How would removing low-correlation, rare filters affect performance? (2) Optimizing ensemble formation. The results show some features (and subspaces) are shared between independently trained DNNs, and some are not. This suggests testing how feature correlation among different DNNs in an ensemble affects ensemble performance. For example, the ``shared cores" of multiple networks could be deduplicated, but the unique features in the tails of their feature sets could be kept. (3) Similarly, one could (a) post-hoc assemble ensembles with greater diversity, or even (b) directly encourage ensemble feature diversity during training. (4) Certain visualization techniques, e.g., deconv \citep{zeiler2014visualizing}, DeepVis \citep{yosinski-2015-ICML-DL-understanding-neural-networks}
, have revealed neurons with multiple functions (e.g. detectors that fire for wheels and faces). The proposed matching methods could reveal more about why these arise. Are these units consistently learned because they are helpful or are they just noisy, imperfect features found in local optima?
(5) Model combination: can multiple models be combined by concatenating their features, deleting those with high overlap, and then fine-tuning?
(6) Apply the analysis to networks with different architectures --- for example, networks with different numbers of layers or different layer sizes --- or networks trained on different subsets of the training data.
(7) Study the correlations of features in the same network, but across training iterations, which could show whether some features are trained early and not changed much later, versus perhaps others being changed in the later stages of fine-tuning. This could lead to complementary
insights on learning dynamics to those reported by \cite{erhan2009difficulty}.
(8) Study whether particular regularization or optimization strategies (e.g., dropout, ordered dropout, path SGD, etc.) increase or decrease the convergent properties of the representations to facilitate different goals (more convergent would be better for data-parallel training, and less convergent would be better for ensemble formation and compilation).

\subsubsection*{Acknowledgments}
\vspace*{-.8em}
The authors are grateful to the NASA Space Technology Research Fellowship (JY) for funding, Wendy Shang for conversations
and initial ideas, Yoshua Bengio for modeling suggestions, and Anh Nguyen and Kilian Weinberger for helpful comments and edits.
This work was supported in part by US Army Research Office W911NF-14-1-0477, NSF grant 1527232.
Jeff Clune was supported by an NSF CAREER award (CAREER: 1453549) and a hardware donation from the NVIDIA Corporation.

{
\bibliography{ref,bibdesk}
\bibliographystyle{iclr2016_conference}             
}
\newpage
\appendix
\renewcommand{\thesection}{S.\arabic{section}}
\renewcommand{\thesubsection}{\thesection.\arabic{subsection}}

\newcommand{\beginsupplementary}{%
        \setcounter{table}{0}
        \renewcommand{\thetable}{S\arabic{table}}%
        \setcounter{figure}{0}
        \renewcommand{\thefigure}{S\arabic{figure}}%
}

\beginsupplementary

\section*{Supplementary Information}

\section{Activation Values: High Correlation vs. Low Correlation}

\begin{figure*}[htbp!]
    \centering
    \begin{subfigure}[t]{0.48\textwidth}
        \centering
        \includegraphics[width=1\textwidth]{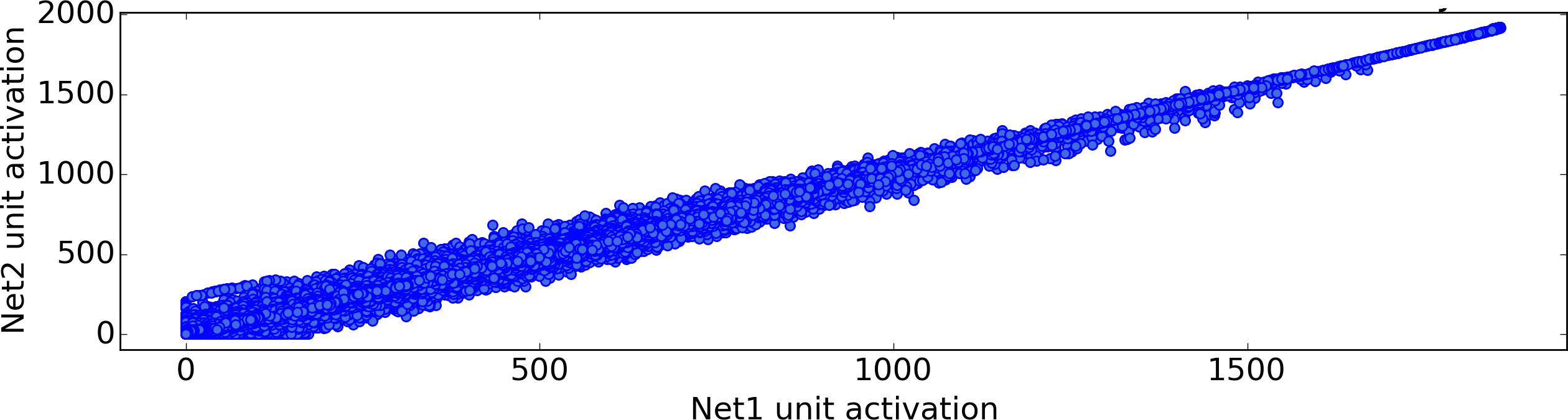}
        \caption{\layer{conv1}, high correlation $c_{1,34,48}^{(1,2)}$: 0.995}
        \end{subfigure}%
    ~ 
    \begin{subfigure}[t]{0.48\textwidth}
        \centering
        \includegraphics[width=1\textwidth]{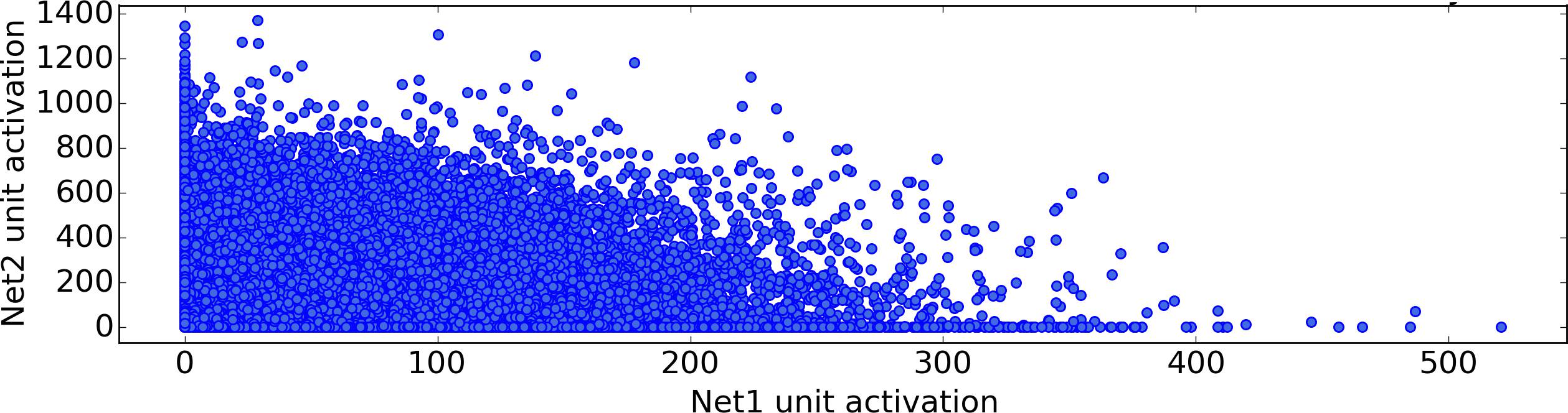}
        \caption{\layer{conv1}, low correlation $c_{1,2,74}^{(1,2)}$: 0.178}
           \end{subfigure}
    \begin{subfigure}[t]{0.48\textwidth}
        \centering
        \includegraphics[width=1\textwidth]{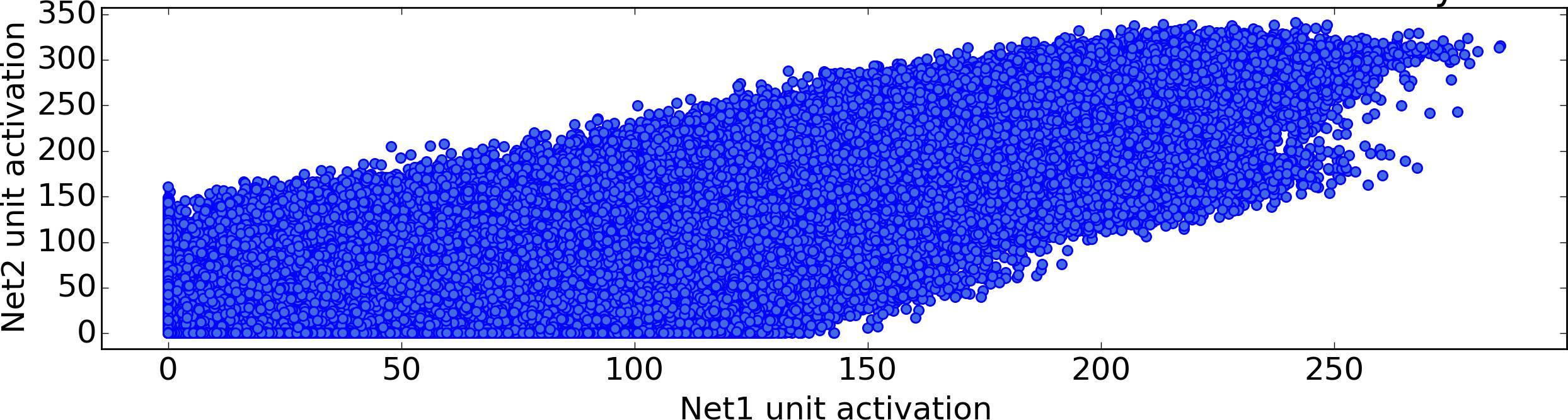}
        \caption{\layer{conv2}, high correlation $c_{2,122,236}^{(1,2)}$: 0.92}       
         \end{subfigure}
    ~
    \begin{subfigure}[t]{0.48\textwidth}
        \centering
        \includegraphics[width=1\textwidth]{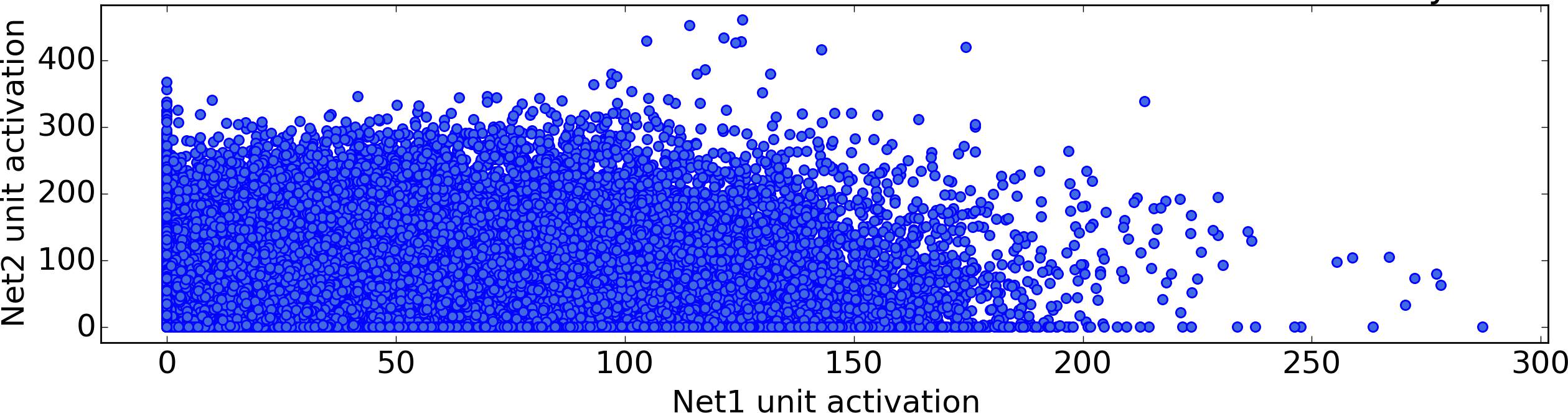}
        \caption{\layer{conv2}, low correlation $c_{2,30,26}^{(1,2)}$: 0.24}
         \end{subfigure}
 \caption{
  Activation values are computed across 5,000 randomly sampled images and all spatial positions ($55 \times 55$ and $27 \times 27$ for layers \layer{conv1} and \layer{conv2}, respectively).
  The joint distributions appear simple enough to suggest that a correlation measure\jmc{, which is linear,} is sufficient to find matching units between networks.
\todo{The more I try to understand this plot, the more confused I get. What are the units of the axes? Originally I thought they were neurons within the net, but now I am pretty sure they are activation levels? But in what unit (i.e. why do the values go from 0 to 2000...what does an activation value of 2000 mean?) Am I right that each blue dot represents the activation of each of the two neurons for one patch, such that there are 5000*55*55 dots for conv1 plots? If so, I think we should say that."
We never explain the notation with 3 numbers as subscripts. Presumably the first number is the layer, the 2nd is the unit in network 1, and the third is the unit in network 2? Probably we should group 2\&3, which are units, differently than 1, which is a layer..or at least explain it. Overall this figure took me a LONG time to parse, and we don't want to burden our reader with that mental exercise. It should be super simple to understand what this means after reading the caption, which it currently is not. Finally, this figure never says what the main takeaway of the figure is. What I am I supposed to learn from it?}}
\figlabel{activation_correlation}
\end{figure*}

\section{Additional Matching Results}

\begin{figure}[t]
  \vspace{-0em}
\begin{center}
  \hspace*{-.08\linewidth}\includegraphics[width=1.05\linewidth]{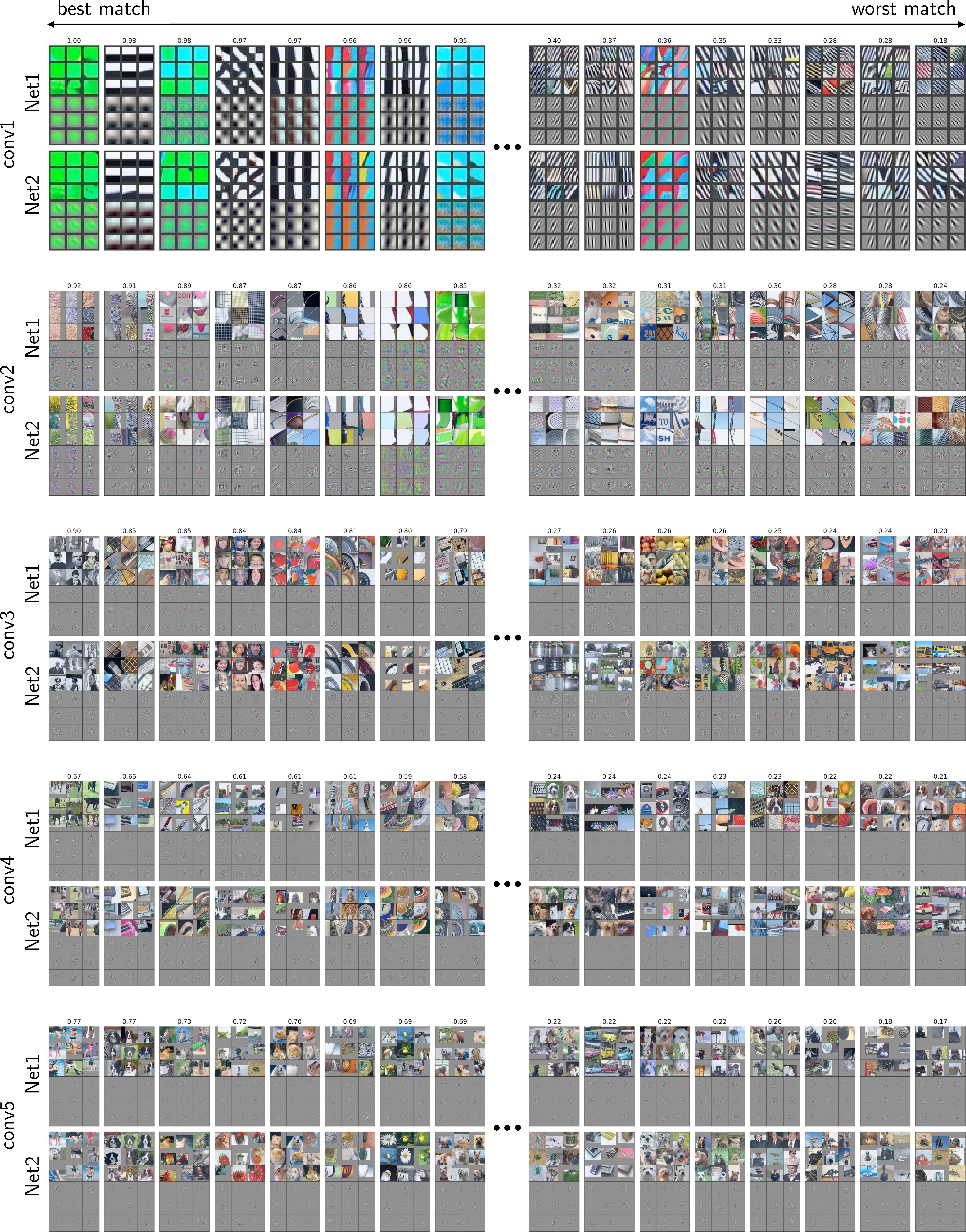}
  \caption{ With assignments chosen by semi-matching, the eight best (highest correlation, left) and eight worst (lowest correlation, right) matched features between \net{1} and \net{2} for the \layer{conv1} through \layer{conv5} layers.
    To visualize the functionality each unit, we plot the nine image patches (in a three by three block) from the validation set that causes the highest activation for that unit and directly beneath that block show the ``deconv'' visualization of each of the nine images. Best view with siginicant zoom in.   
  }
  \figlabel{match_ims_top_bot_conv4_conv5}
\end{center}
\end{figure}

\figref{match_vs_max_all} shows additional results of  comparison of assignments produced by semi-matching and matching methods in \layer{conv2} -- \layer{conv5}. For each unit, both the
 semi-matching and matching are found, then the units are sorted in order of decreasing semi-matching value and both correlation values are plotted.

\begin{figure*}[htbp!]
    \centering
    \begin{subfigure}[t]{0.48\textwidth}
        \centering
        \includegraphics[width=1\textwidth]{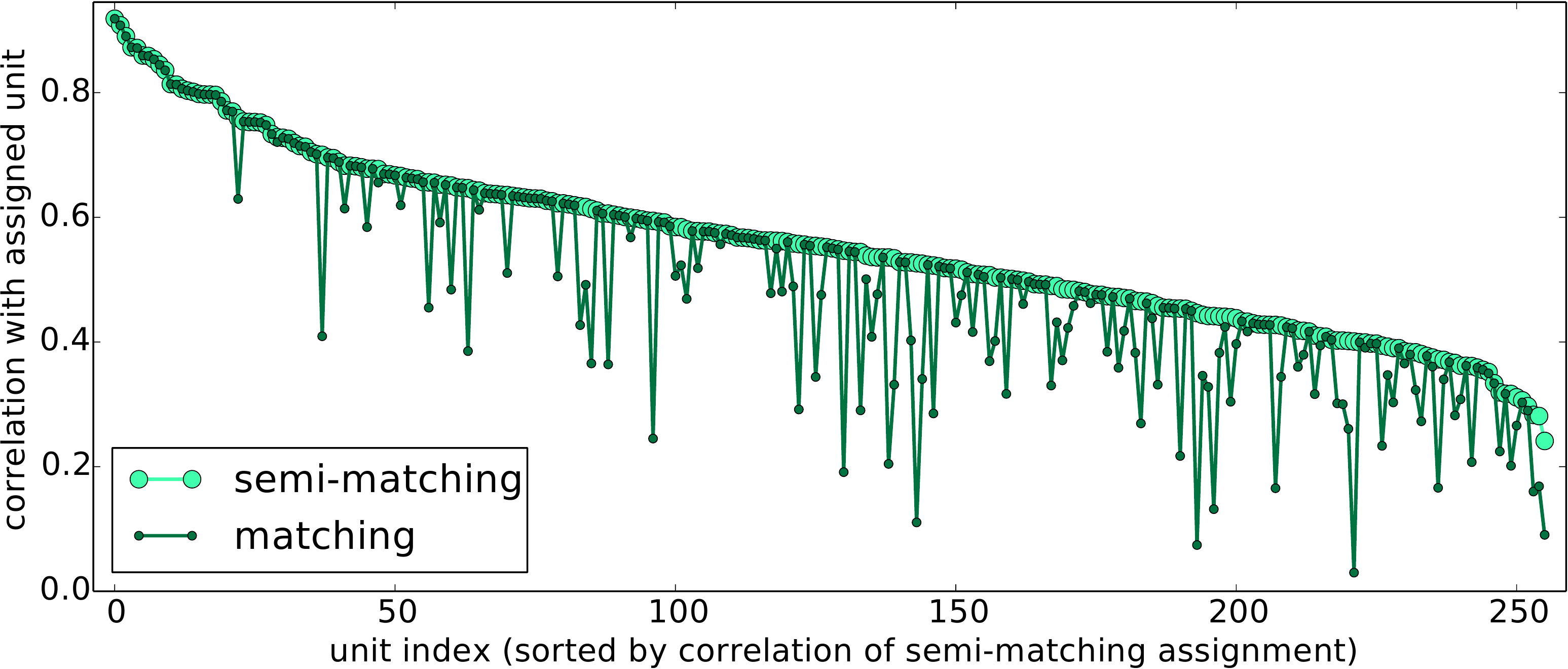}
        \caption{\layer{conv2}}
        \end{subfigure}%
    ~ 
    \begin{subfigure}[t]{0.48\textwidth}
        \centering
        \includegraphics[width=1\textwidth]{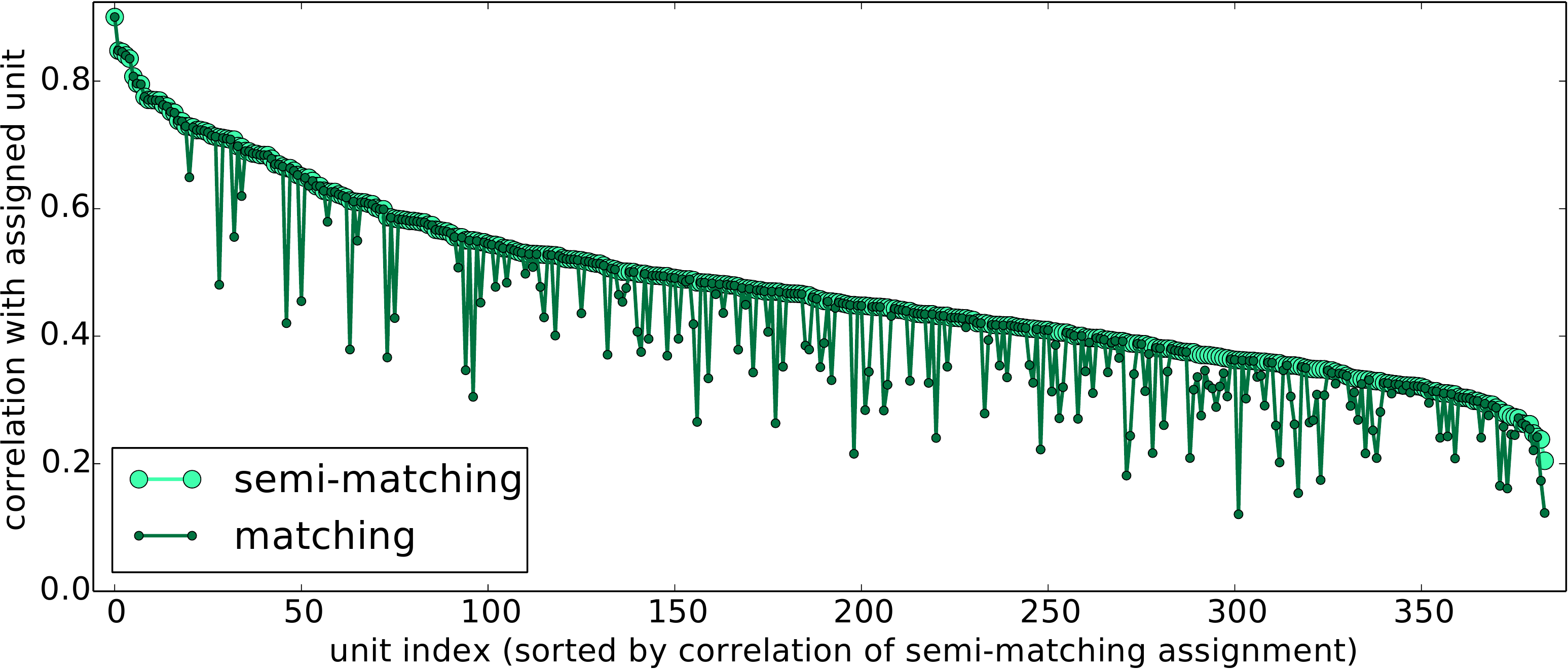}
        \caption{\layer{conv3}}
           \end{subfigure}
    \begin{subfigure}[t]{0.48\textwidth}
        \centering
        \includegraphics[width=1\textwidth]{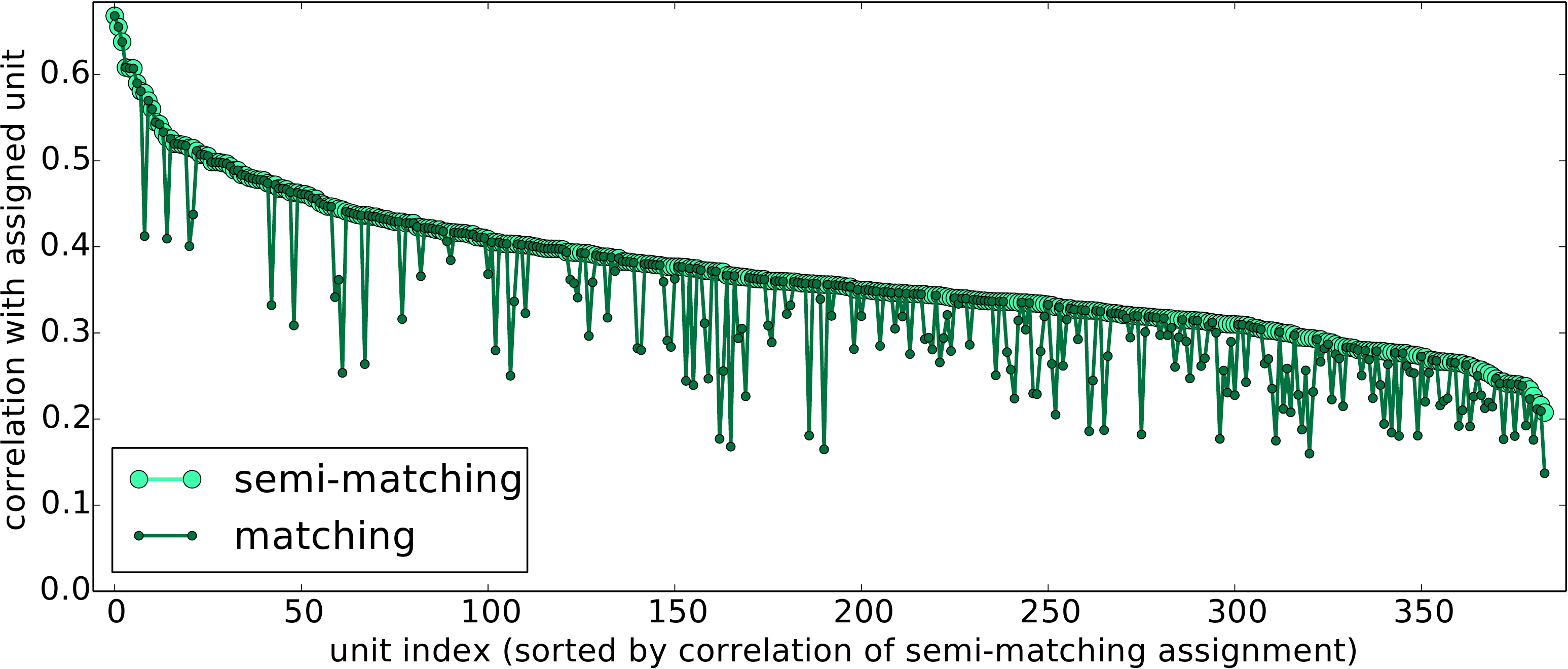}
        \caption{\layer{conv4}}       
         \end{subfigure}
    ~
    \begin{subfigure}[t]{0.48\textwidth}
        \centering
        \includegraphics[width=1\textwidth]{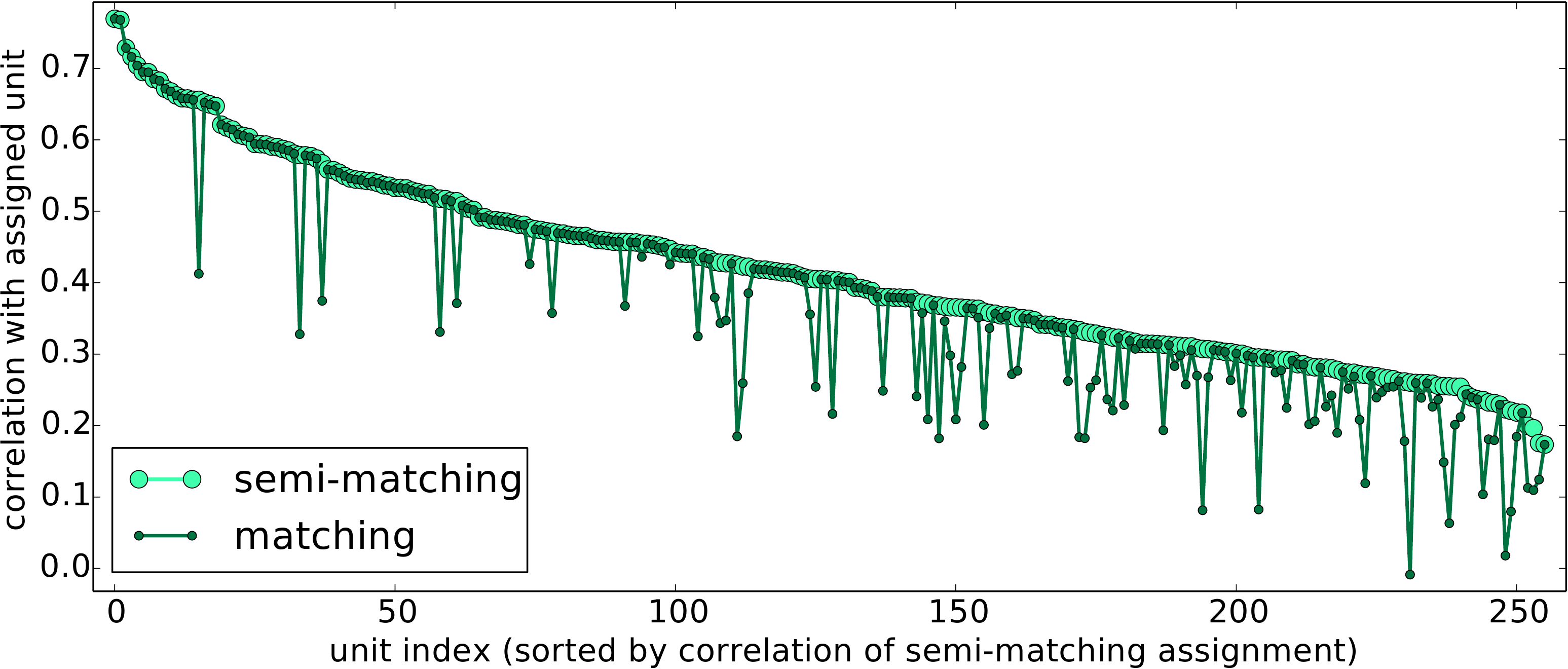}
        \caption{\layer{conv5}}
         \end{subfigure}
 \caption{
  Correlations between units in \layer{conv2} - \layer{conv5} layers of Net1 and their paired units in Net2, where pairings are made via semi-matching (large light green circles) or matching (small dark green dots).}
\figlabel{match_vs_max_all}
\end{figure*}





\newpage

\begin{figure}[htbp]
\begin{center}
  \includegraphics[width=1\linewidth]{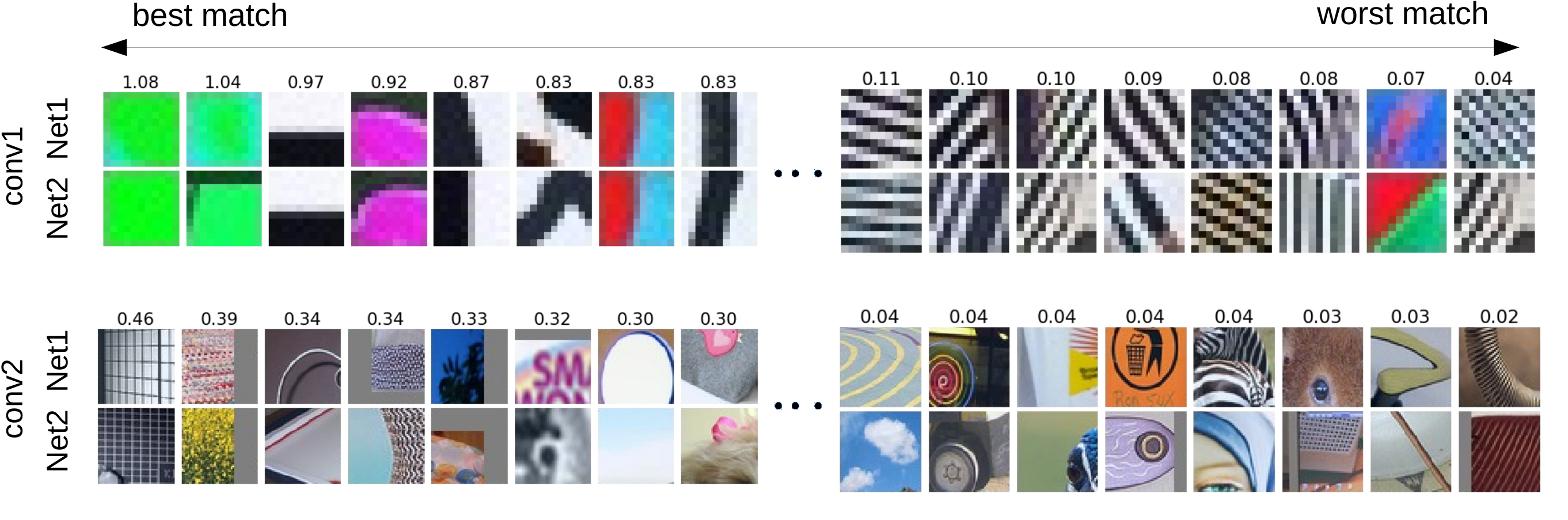}
  \caption{
   The eight best (highest mutual information, left) and eight worst (lowest mutual information, right) features in the semi-matching between \net{1} and \net{2} for the \layer{conv1} and \layer{conv2} layers.
   \later{Investigate how the worse matches are so bad, e..g .04, when the correlation is .18? A non-zero correlation implies positive MI.}
  }
  \figlabel{match_mi_top_bot}
\end{center}
\end{figure}


\section{Does Relaxing the One-to-One Constraint to Find Many-to-Many Groupings Reveal More Similarities Between what Different Networks Learn?}
\seclabel{spectral}


\todo{reword the below sentence to be dissimilar to the sparse prediction section}
Since the preceding sections have shown that neurons may not necessarily correspond via a globally
consistent one-to-one matching between networks, we now seek to find
many-to-many matchings between networks 
using a spectral clustering approach.



\subsection{Neuron Similarity Graphs}
\seclabel{sim_graph}

We define three types of similarity graphs based on the correlation matrices obtained above. 

{\bf Single-net neuron similarity graphs.} Given a fully trained DNN $X$ and a specified layer $l$, we first construct the {\em single-net neuron similarity graph} $G_{X,l}=(V,E)$. Each vertex $v_p$ in this graph represents a unit $p$ 
 in layer $l$. Two vertices are connected by an edge of weight $c_{pq}$ if the correlation value $c_{pq}$ in the self-correlation matrix $\mathrm{corr}(X_l,X_l)$ between unit $p$ and unit $q$ is greater than a certain threshold $\tau$. 

{\bf Between-net neuron similarity graphs.}  Given a pair of fully trained DNNs $X$ and $Y$, the {\em between-net neuron similarity graph} $G_{XY,l}=(V,E)$ can be constructed in a similar manner. Note that $G_{XY,l}$ is a bipartite graph and  contains twice as many vertices as that in $G_{X,l}$ since it incorporates units from both networks.  Two vertices are connected by an edge of weight $c_{pq}$ if the correlation value $c_{pq}$ in the between-net correlation matrix $\mathrm{corr}(X_l,Y_l)$ between unit $p$ in $X$ and unit $q$ in $Y$ is greater than a certain threshold $\tau$.

{\bf Combined similarity graphs.} The problem of matching neurons in different networks can now be reformulated as finding a partition in the combined neuron similarity graphs 
$G_{X+Y,l}=$ $G_{X,l}$ $+$ $G_{Y,l} + G_{XY,l}$, such that the edges between different groups have very low weights and the edges within a group have relatively high weights. 

\subsection{Spectral Clustering and Metrics for Neuron Clusters}

We define three types of similarity graphs based on the correlation matrices obtained above (see \secref{sim_graph} for definition). 
Define $W_l \in \mathbb{R}^{2\mathcal{S}_l \times 2\mathcal{S}_l }$ to be the combined correlation matrices between two DNNs, $X$ and $Y$ in layer $l$, where $w_{jk}$ is the entry at $j$th row and $k$th column of that matrix. $\mathcal{S}_l$ is the number of channels (units) in layer $l$. $W_l$ is given by
$$
W_l = \begin{bmatrix}
\mathrm{corr}(X_l,X_l) & ~\mathrm{corr}(X_l,Y_l) \\
~~\mathrm{corr}(X_l,Y_l)^\top & \mathrm{corr}(Y_l,Y_l)
\end{bmatrix}.
$$ 
The unnormalized  Laplacian matrix is defined as $L_l = D_l - W_l$,
where the degree matrix $D_l$ is the diagonal matrix with the degrees $d_j = \sum_{k=1}^{2\mathcal{S}_l} w_{jk}$. The unnormalized Laplacian matrix
and its eigenvalues and eigenvectors can be used to effectively embed points in a lower-dimension representation without losing too information about spatial relationships.
If neuron clusters can be identified, then the Laplacian $L_l$  is approximately block-diagonal, with each block corresponding a cluster. Assuming there are $k$ clusters in the graph, spectral clustering would then take the first $k$ eigenvectors\footnotemark, $\mathcal{U} \in \mathbb{R}^{2\mathcal{S}_l\times k}$, corresponding to the $k$ smallest
eigenvalues, and partition neurons in the eigenspace with the $k$-means algorithm.

\footnotetext{In practice, when the number of clusters is unknown, the best value of $k$ to choose is where where the eigenvalue shows a relatively  abrupt change.}



\seclabel{clustering}

\subsection{Spectral Clustering Results}

We use \net{1} and \net{2} as an example for showing the results of matching neurons between DNNs. \figref{similarity_sup} shows the permuted combined correlation matrix after apply the spectral clustering algorithm for \layer{conv1} -- \layer{conv5}. \figref{conv1_visual} displays 12 neuron clusters with high between-net similarity measurement in \layer{conv1} layer, and \figref{conv2_visual} displays the top 8 neuron clusters with highest between-net similarity measurement in \layer{conv2} layer.   The matching results imply that there exists many-to-many correspondence of the feature maps between two fully trained networks with different random initializations, and the number of neurons learning the same feature can be different between networks. For example, the four units of \{89, 90, 134, 226\} in \net{1} and three units of  \{2, 39, 83\} in \net{2} are learning the features about green objects.

\begin{figure*}[htbp]
    \centering
    \begin{subfigure}[t]{0.38\textwidth}
        \centering
        \includegraphics[width=1\textwidth]{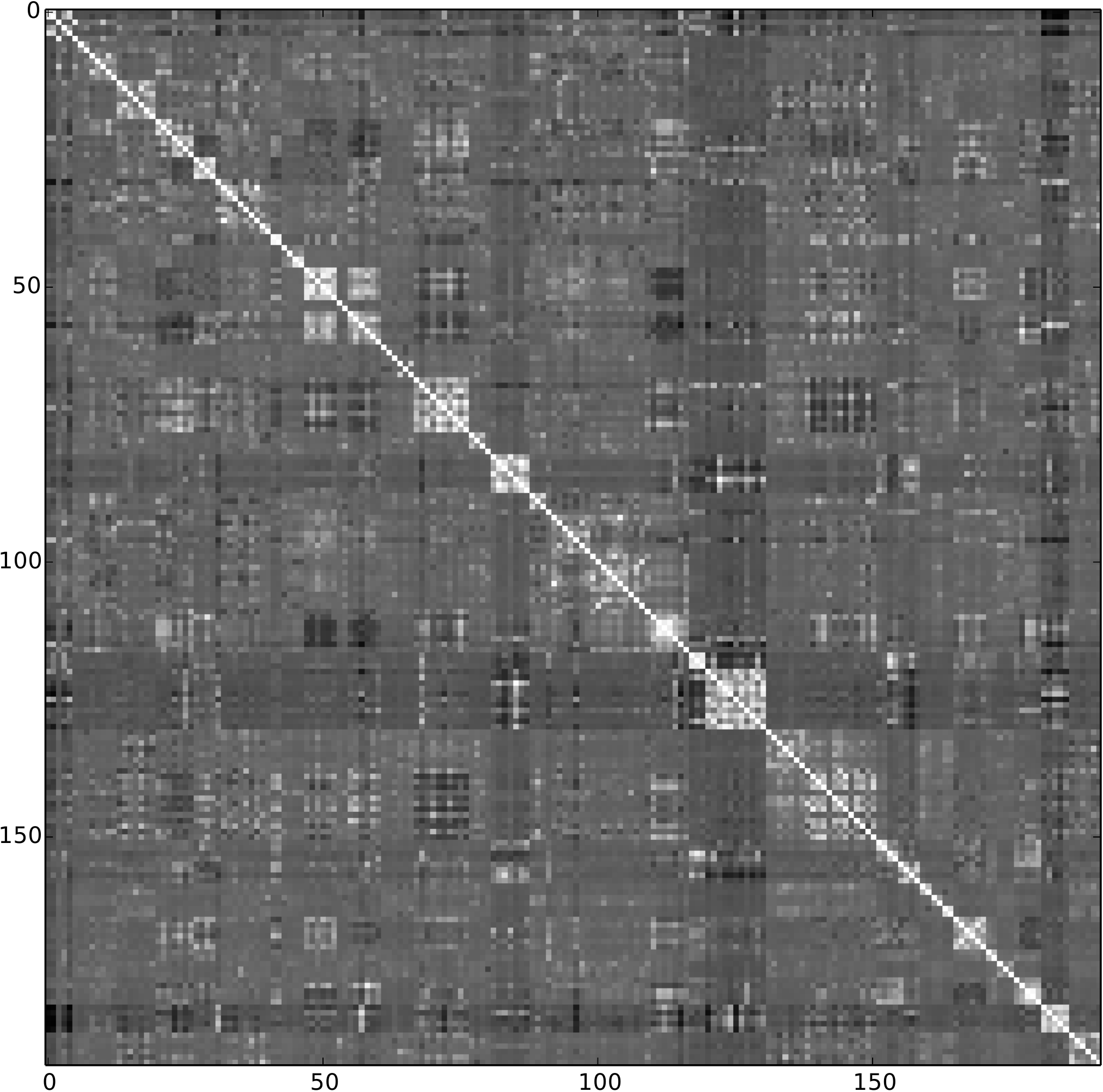}
        \caption{\layer{conv1}}
        \end{subfigure}%
    ~ 
    \begin{subfigure}[t]{0.38\textwidth}
        \centering
        \includegraphics[width=1\textwidth]{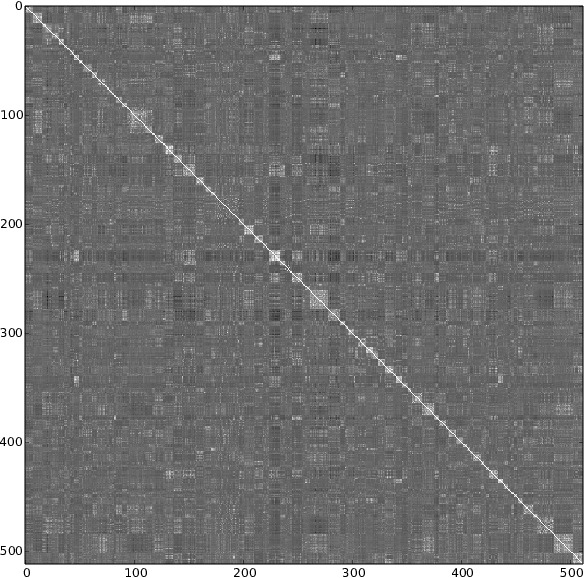}
        \caption{\layer{conv2}}
           \end{subfigure}
   
    \begin{subfigure}[t]{0.38\textwidth}
        \centering
        \includegraphics[width=1\textwidth]{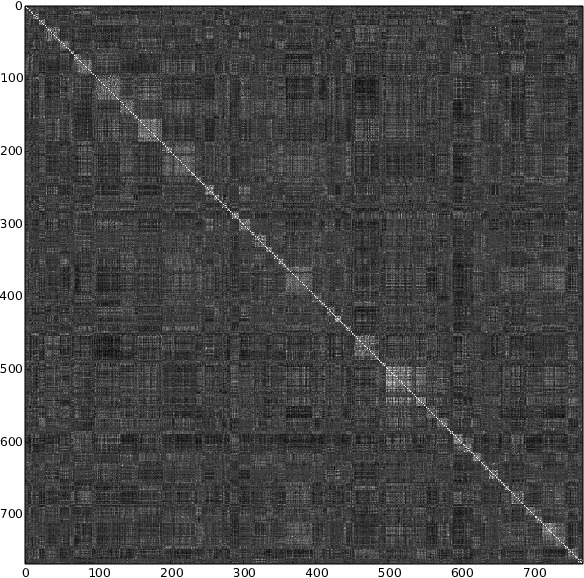}
        \caption{\layer{conv3}}       
         \end{subfigure}
       ~
                  \begin{subfigure}[t]{0.38\textwidth}
        \centering
        \includegraphics[width=1\textwidth]{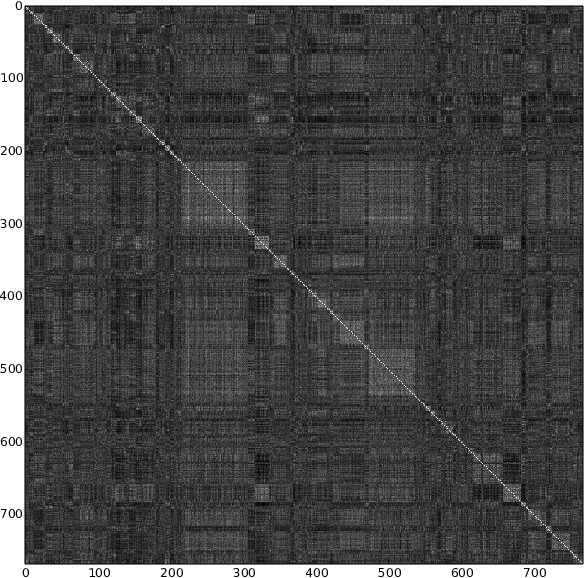}
        \caption{\layer{conv4}}
         \end{subfigure}
         
                      \begin{subfigure}[t]{0.38\textwidth}
        \centering
        \includegraphics[width=1\textwidth]{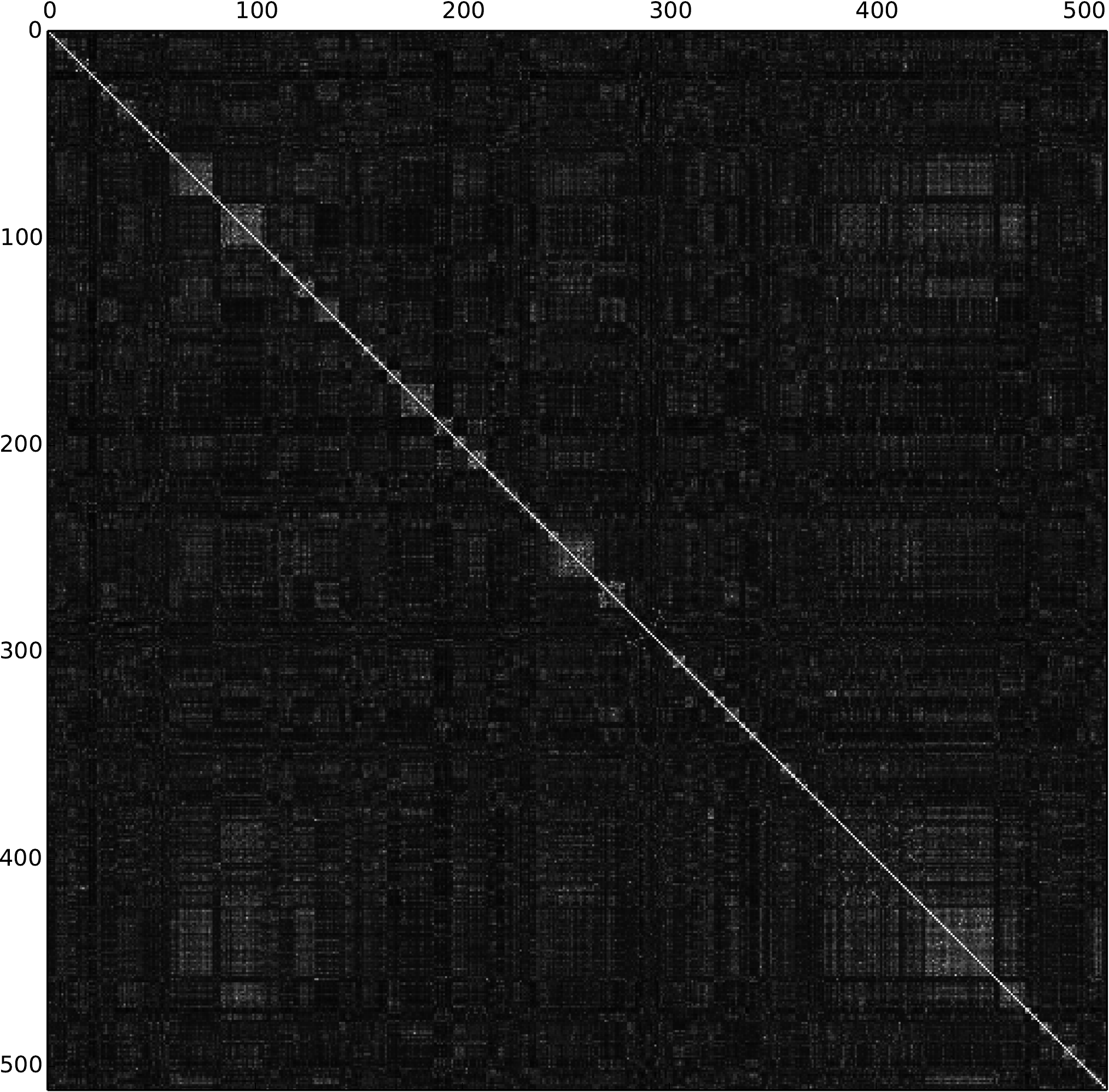}
        \caption{\layer{conv5}}
         \end{subfigure}

    \caption{The permuted combined correlation matrix after apply spectral clustering method (\layer{conv1} -- \layer{conv5}). The diagonal block structure represents the groups of neurons that are clustered together. The value of $k$ adopted for these five layers are: \{40,100,100,100,100\}, which is consistent with the parameter setting for other experiments in this paper.} 
    \figlabel{similarity_sup}
\end{figure*}

\begin{figure}[htbp]
\begin{center} \includegraphics[width=1\textwidth]{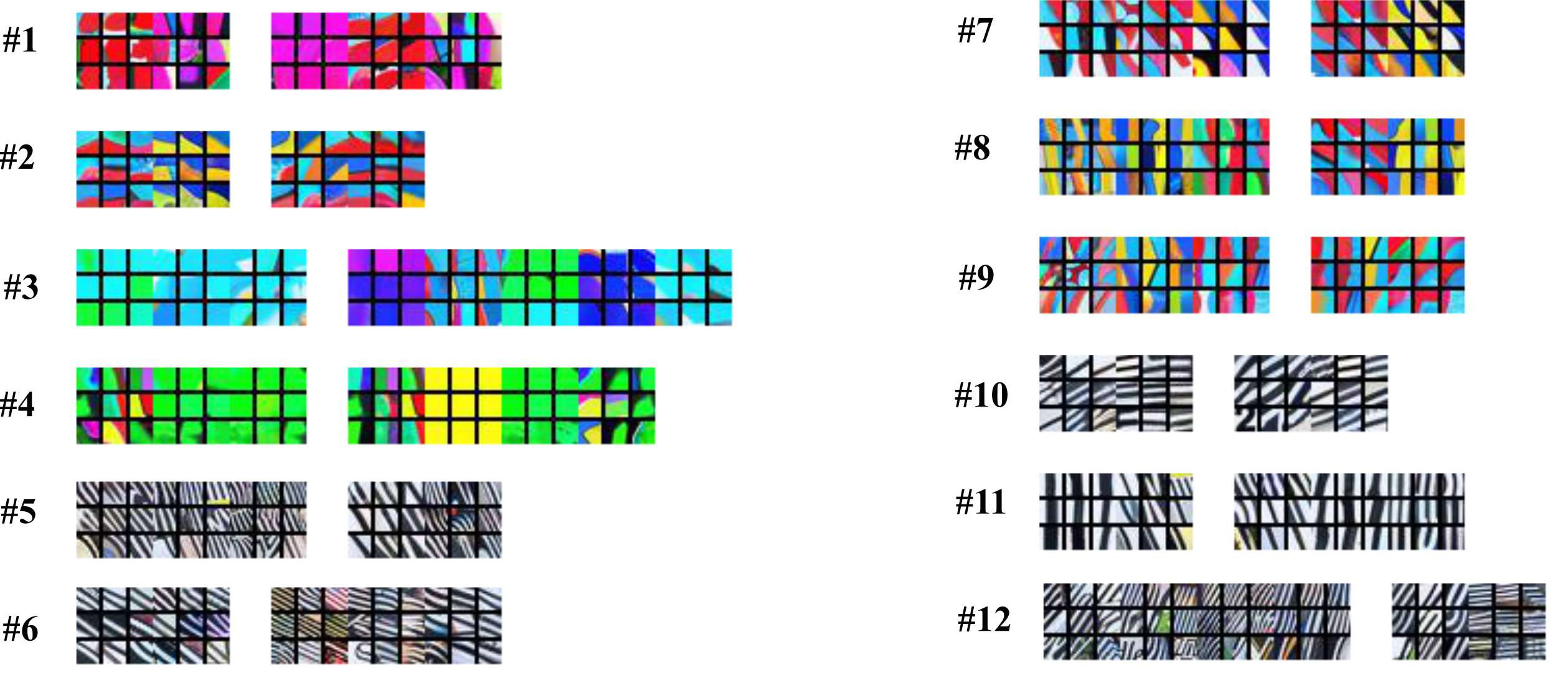} \end{center}
\vspace{-1em}
\caption{The neuron matchings between two DNNs  (\net{1} and \net{2}) in \layer{conv1} layer.  Here we display the 12 neuron clusters with relatively high between-net similarity measurement. Each labeled half-row corresponds to one cluster, where the filter visualizations for neurons from \net{1} and \net{2} are separated by white space slot. The matching results imply that there exists many-to-many correspondence of the feature maps between two fully trained networks with different random initializations. For instance, in cluster \#6, neurons from \net{1} and \net{2} are both learning 135$^{\circ}$ diagonal edges; and neurons in cluster \#10 and \#12 are learning 45$^{\circ}$ diagonal edges. }
\figlabel{conv1_visual}
\end{figure}

\begin{figure}[t]
  \begin{center}
    \includegraphics[width=1\textwidth]{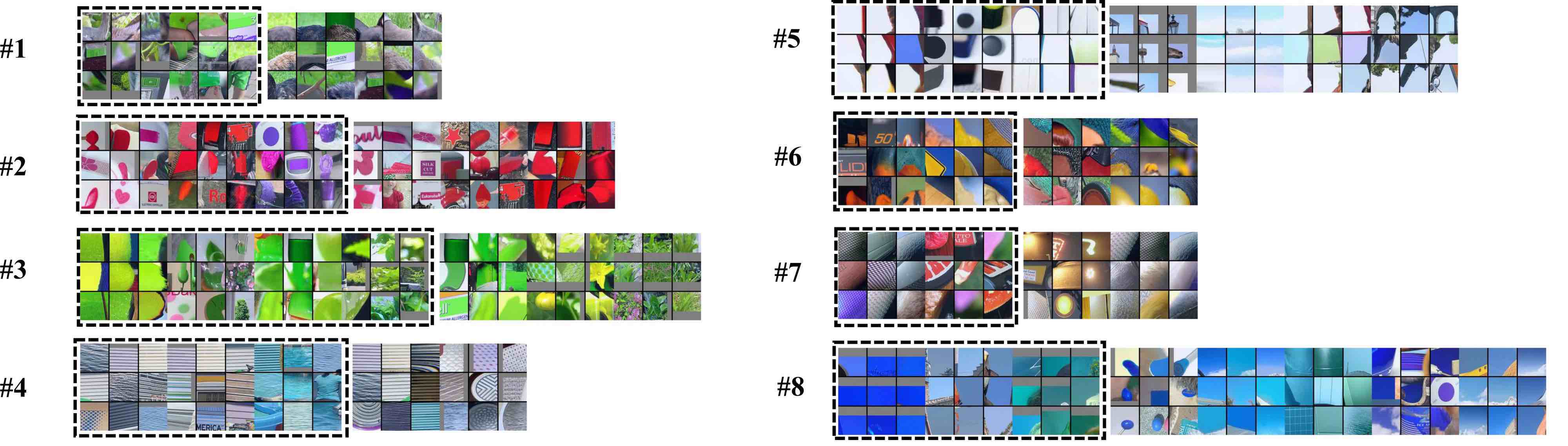}
  \end{center}
\caption{The neuron matchings between two DNNs: \net{1} and \net{2}. Each of the 3 $\times$ 3 block displays the top 9 image patches that cause the highest activations to each neuron. Each labeled half-row corresponds to one cluster, where the filter visualizations with dashed boxes represent neurons from \net{1} and those without are from \net{2}. For example, there are 7 neurons learning similar features in cluster \#3, where the left four neurons are in \net{1} and the right three are from \net{2}.  
Best viewed in electronic form with zoom.}
\figlabel{conv2_visual}
\vspace{-1em}
\end{figure}

We also computed and visualized the matching  neurons in other layers as well. The results of \layer{conv1} are shown in \figref{conv1_visual}. As noted earlier, the \layer{conv1} layer tends to learn more general features like Gabor filters (edge detectors) and blobs of color. Our approach finds many matching Gabor filters (e.g., clusters \#5, \#6, \#10, \#11 and \#12), and also some matching color blobs (e.g., clusters \#1,  \#3 and \#4).

\vspace*{-.5em}
\subsection{Hierarchical Spectral Clustering Results}
\vspace*{-.5em}

Due to the stochastic effects of randomly initializing centroids in $k$-means clustering, some of the initial clusters contains more neurons than others. To get more fine-grained cluster structure, we recurrently  apply $k$-means clustering on any clusters with size $> 2 \alpha \cdot\mathcal{S}_l$, where $\alpha$ is a tunable parameter for adjusting the maximum size of the leaf clusters. \figref{hierarchy} shows the partial hierarchical structure of neuron matchings in the \layer{conv2} layer. The cluster at the root of the tree is a first-level cluster that contains many similar units from both DNNs. Here we adopt $\alpha=0.025$ for the \layer{conv2} layer, resulting in a hierarchical neuron cluster tree structure with leaf clusters containing less than 6 neurons from each network. The bold box of each subcluster contains neurons from \net{1} and the remaining neurons are from \net{2}. For example, in subcluster \#3, which shows \layer{conv2} features, units \{62, 137, 148\} from \net{1} learned similar features as units \{33, 64, 230\} from \net{2}, namely, red and magenta objects.

\begin{figure}[h]
\begin{center} \includegraphics[width=1\textwidth]{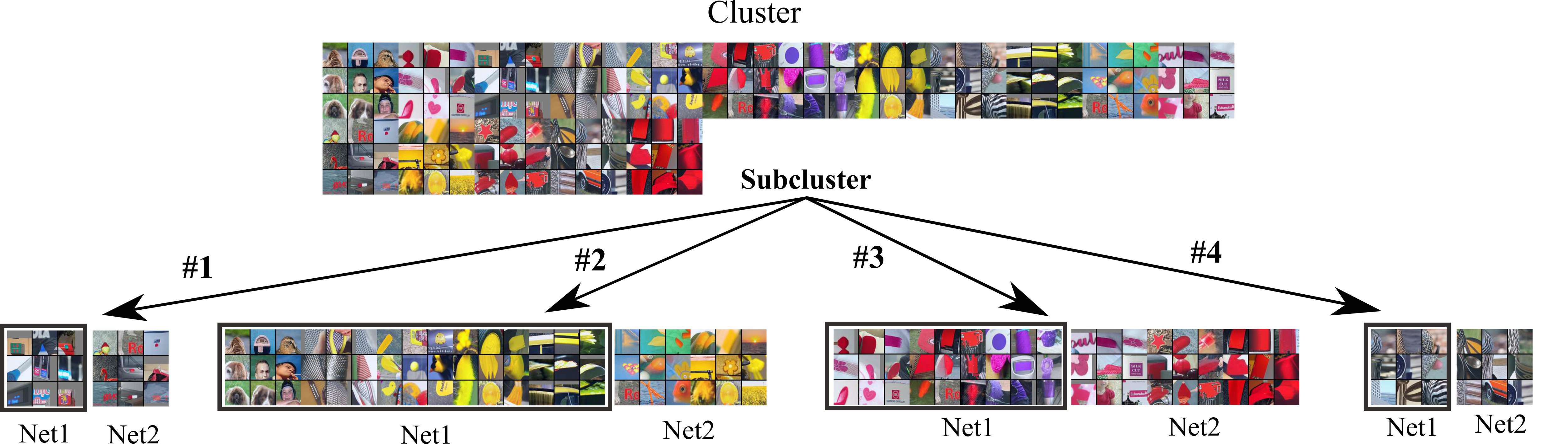} \end{center}
\caption{The hierarchical structure of neuron matchings between two DNNs: \net{1} and \net{2} (\layer{conv2} layer). The initial clusters are obtained using spectral clustering with the number of clusters $k=100$ and threshold $\tau= 0.2$. }
\figlabel{hierarchy}
\end{figure}



\subsection{Metrics for Neuron Clusters}
\seclabel{similarity}
%

Here we introduce two metrics for quantifying the similarity among neurons grouped together after applying the clustering algorithm above.
\newcommand{\eqlabel}[1]{\textrm{#1}}
\begin{eqnarray*}
\eqlabel{Between-net similarity:}\hspace{2em} \mathrm{Sim}_{X_l\rightarrow  Y_l} &=& ( \sum_{p=1}^{\mathcal{S}_l}\sum_{q=1}^{\mathcal{S}_l}\mathrm{corr}(X_l,Y_l)_{pq} ) / \mathcal{S}_l^2   \\
\eqlabel{Within-net similarity:}\hspace{2em} \mathrm{Sim}_{X_l,Y_l}& =&  (\mathrm{Sim}_{X_l\rightarrow  X_l} + \mathrm{Sim}_{Y_l\rightarrow  Y_l}) / 2
\end{eqnarray*}

We further performed experiments in quantifying the similarity among neurons that are clustered together.  \figref{similarity} shows the between-net and within-net similarity measurement for \layer{conv1} -- \layer{conv5}. The value of $k$ for initial clustering is set to be 40 for \layer{conv1} layer and 100 for all the other layers. In our experiments, the number of final clusters obtained after further hierarchical branching is \{43, 113,  130, 155, 131\}. The tail in those curves with value 0 is due to the non-existence of between-net similarity for the clusters containing neurons from only one of the two DNNs. To better capture the distribution of non-zero similarity values, we leave out the tail after 100 in the plot for \layer{conv3} - \layer{conv5} layers.

\begin{figure*}[htbp!]
    \centering
    \begin{subfigure}[t]{0.32\textwidth}
        \centering
        \includegraphics[width=1\textwidth]{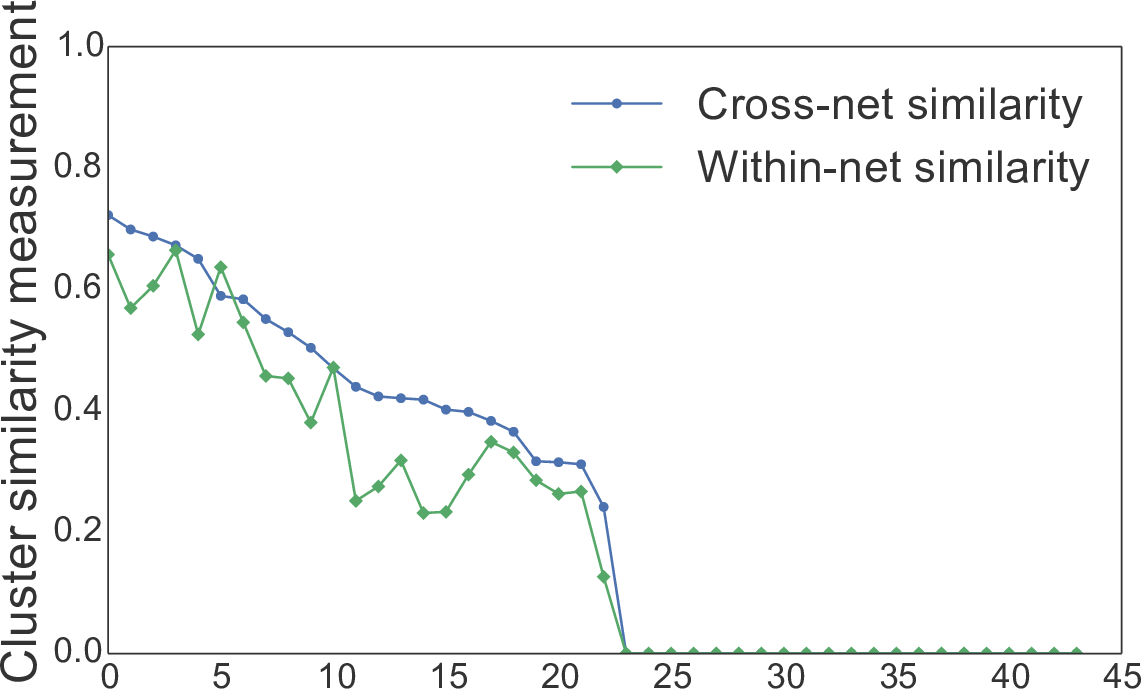}
        \caption{\layer{conv1}}
        \end{subfigure}%
    ~ 
    \begin{subfigure}[t]{0.32\textwidth}
        \centering
        \includegraphics[width=1\textwidth]{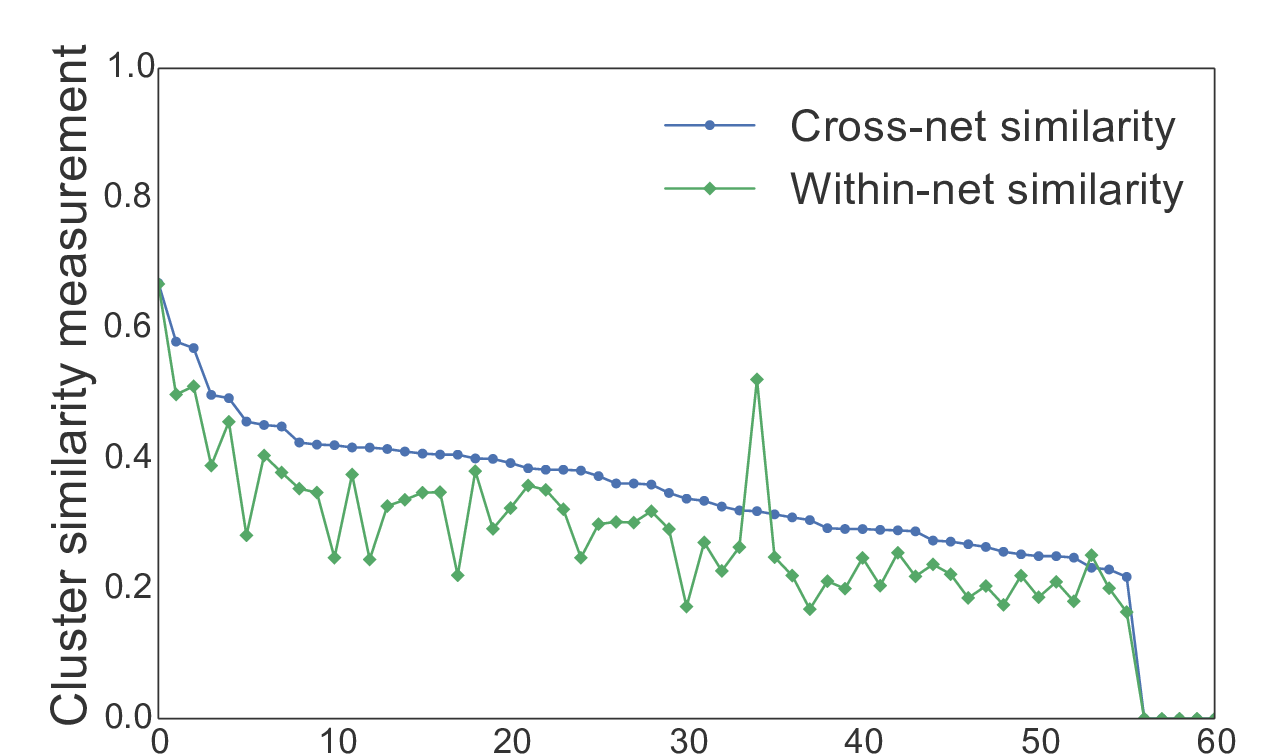}
        \caption{\layer{conv2}}
           \end{subfigure}
   ~
    \begin{subfigure}[t]{0.32\textwidth}
        \centering
        \includegraphics[width=1\textwidth]{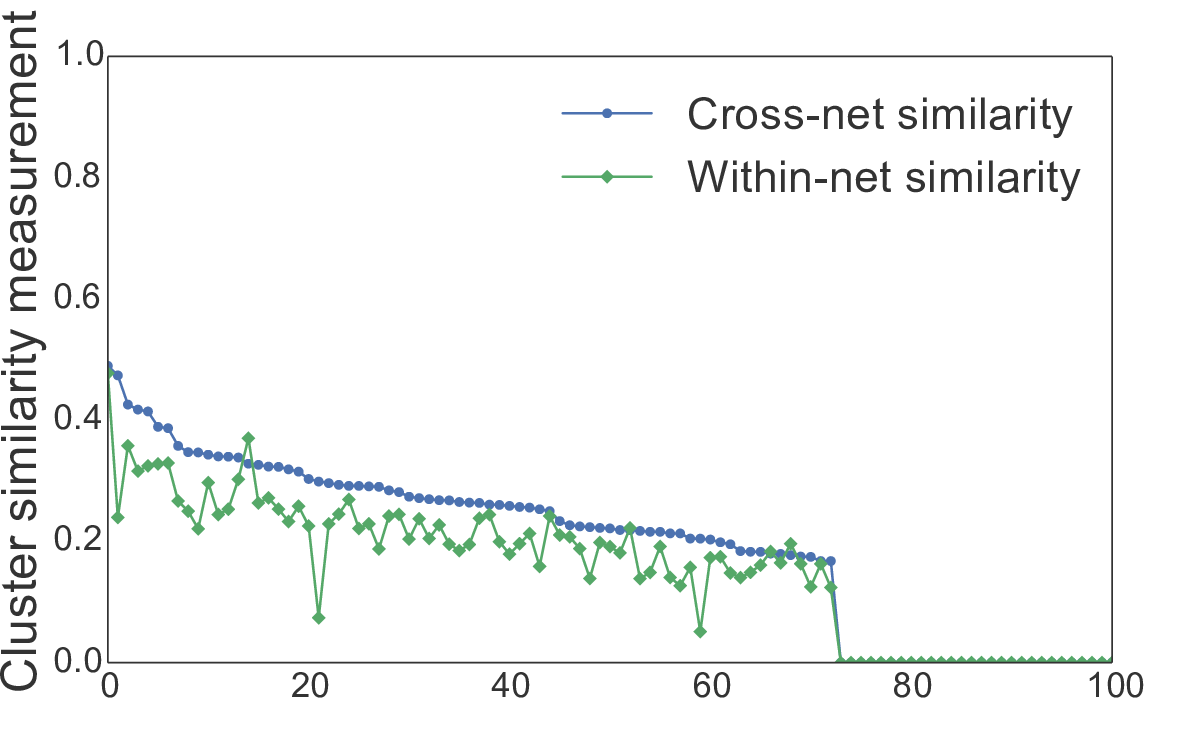}
        \caption{\layer{conv3}}
        
         \end{subfigure}

                  \begin{subfigure}[t]{0.32\textwidth}
        \centering
        \includegraphics[width=1\textwidth]{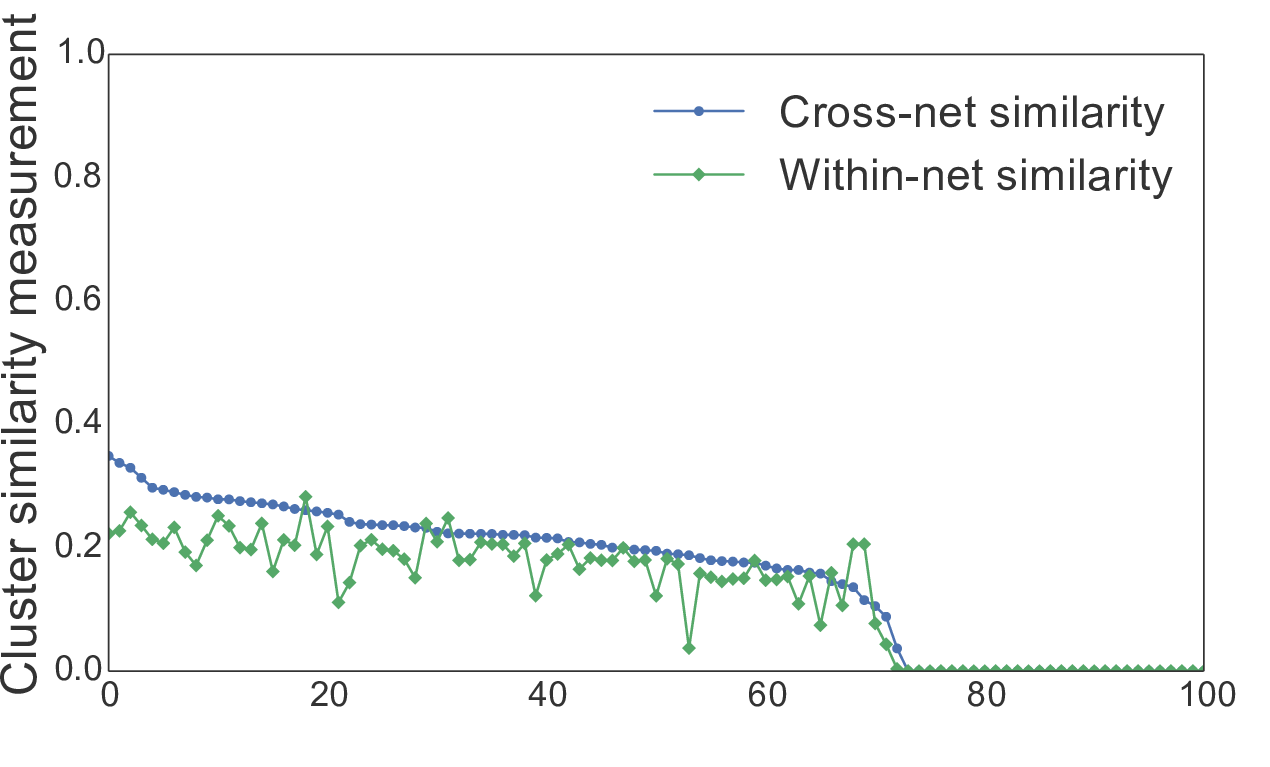}
        \caption{\layer{conv4}}
         \end{subfigure}
                      \begin{subfigure}[t]{0.32\textwidth}
        \centering
        \includegraphics[width=1\textwidth]{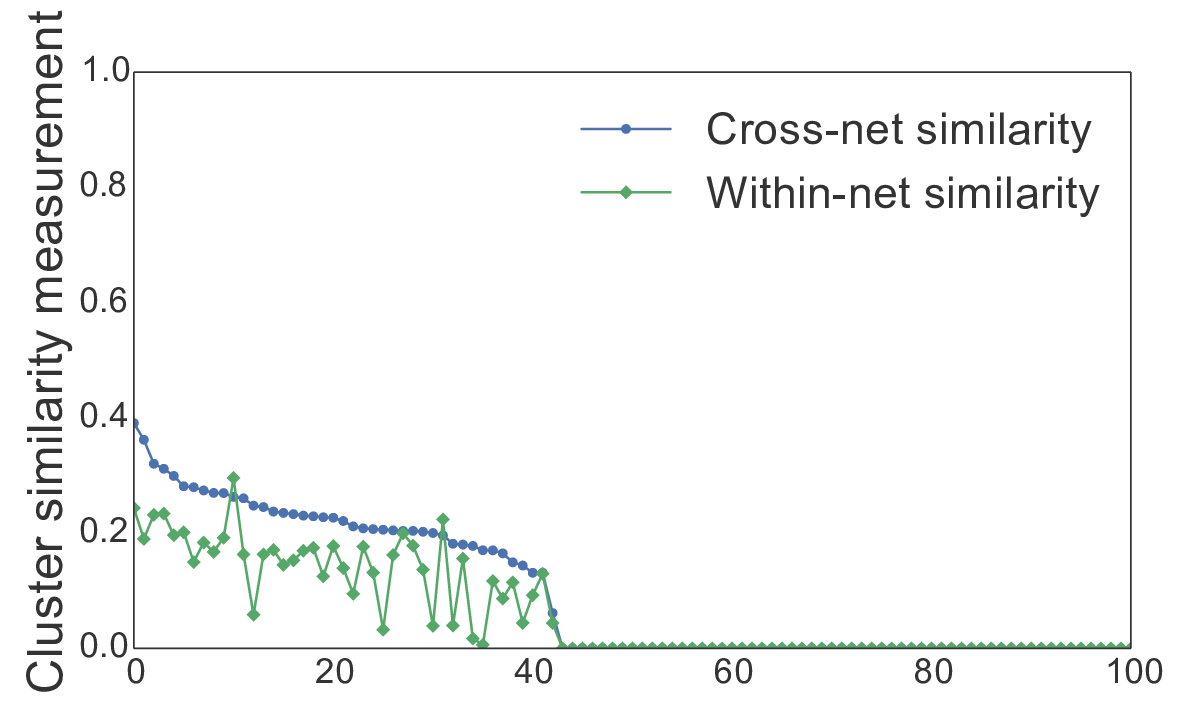}
        \caption{\layer{conv5}}
         \end{subfigure}
                      \caption{The distribution of between-net and within-net similarity measurement after clustering neurons (\layer{conv1} -- \layer{conv5}). The x-axis represents obtained clusters, which is reshuffled according to the sorted between-net similarity value.
                        \todo{export smaller versions of these pdfs and replace ugly png hack!}
 } 

    \figlabel{similarity}
\end{figure*}

\vspace*{-.5em}
\section{Comparing Average Neural Activations within and between Networks}
\seclabel{means}
\vspace*{-.5em}


The first layer of networks trained on natural images (here the \layer{conv1} layer) tends to learn channels matching patterns similar to Gabor filters (oriented edge filters) and blobs of color.
As shown in Figures~\ref{fig:means_nets1234}, \ref{fig:highest_lowest_conv1}, and \ref{fig:highest_lowest_conv2}, there are certain systematic biases in the relative magnitudes of the activations of the different channels of the first layer. Responses of the low frequency filters have much higher magnitude than that of the high frequency filters. This phenomenon is likely a consequence of the $1/f$ power spectrum of natural images in which, on average, low spatial frequencies tend to contain higher energy (because they are more common) than high spatial frequencies.

\begin{figure}[htbp]
\begin{center}
 \includegraphics[width=1\linewidth]{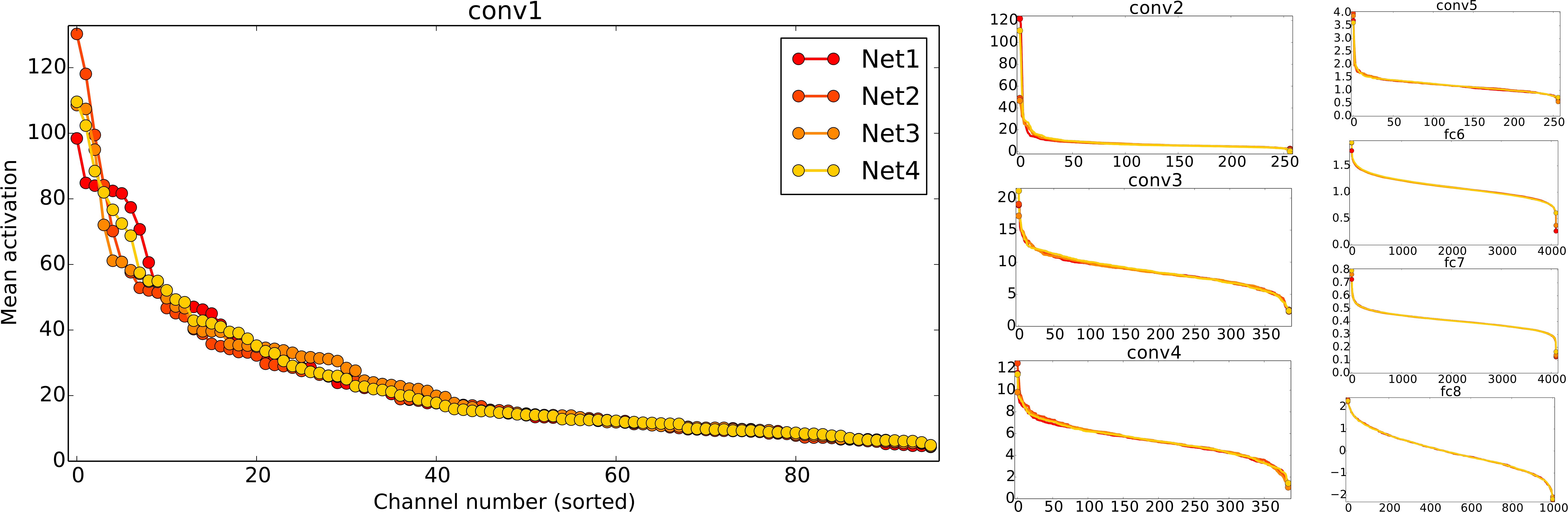}
 \caption{The average activation values of each unit on all layers of \net{1} -- \net{4}. A couple salient effects are observable.
   First, in a given network, average activations vary widely within each layer. While most activations fall within a relatively narrow band (the middle of each plot), a rare few highly active units have one or two orders of magnitude higher average output than the least active.
   Second, the overall distribution of activation values is similar across networks. However, also note that the max single activation does vary across networks in some cases, e.g. on the \layer{conv2} layer by a factor of two between networks.
   For clarity, on layers other than \layer{conv1} circle markers are shown only at the line endpoints.
   \todo{Jason: make bigger}
   \maybe{try more different colors.}
   \maybe{replot with match order, perhaps elsewhere in the paper.}
   \maybe{show which units these are!}
   \maybe{text annotate endpoints with values?}
   \maybe{include dotted gray CDF of gaussian plot}
 }
 \figlabel{means_nets1234}
\end{center}
\end{figure}

In \figref{means_nets1234} we show the mean activations for each unit of four networks, plotted in sorted order from highest to lowest. First and most saliently, we see a pattern of widely varying mean activation values across units, with a gap between the most active and least active units of one or two orders of magnitude (depending on the layer).
Second, we observe a
rough overall correspondence in the spectrum of activations between the networks.
However, the correspondence is not perfect: although much of the spectrum matches well, the most active filters converged to solutions of somewhat different magnitudes.
For example, the average activation value of the filter on \layer{conv2} with the highest average activation varies between 49 to 120
over the four networks; the range for \layer{conv1} was 98 to 130.\footnote{Recall that the units use rectified linear activation functions, so the activation magnitude is unbounded. The max activation over all channels and all spatial positions of the first layer is often over 2000.}
This effect is more interesting considering that all filters were learned with constant weight decay, which pushes all individual filter weights and biases (and thus subsequent activations) toward zero with the same force.
\jmc{why does that make it more surprising? Update: you addressed this by saying that the weight decay is consistent for all networks, but I don't see how that helps explain why we think this result is surprising in light of weight decay.}
\todo{note that the variance of the 10th and 20th units are very small between networks, but the single most active varies a lot between networks!}
 
\figref{highest_lowest_conv1} shows the \layer{conv1} units with the highest and lowest activations for each of the four networks.
As mentioned earlier (and as expected), filters for lower spatial frequencies have higher average activation, and vice versa. What is surprising \todo{soften ``surprising''. ``It could have been this way or that way; here we show that it's this way''.  ``This gets to the heart of the question of to what extent is training convergent. This answers it to some extent''} is the relative lack of ordering between the four networks. For example, the top two most active filters in \net{1} respond to constant color regions of black or light blue, whereas none of the top eight filters in \net{2} or \net{3} respond to such patterns. One might have thought that whatever influence from the dataset caused the largest filters to be black and blue in the first network would have caused similar constant color patches to dominate the other networks, but we did not observe such consistency.
\anremoved{What does it tell us? I think it means that random initialization causes NetX and NetY to learn e.g. the same cyan color detectors but with different responsiveness.}
\jmcremoved{I agree that we should either speculate why this may occur, or, probably better/safer, admit that we do not know why such heterogeneity exists between networks.}
\jmc{I still don't think we have addressed Anh's point or mine. What is the importance/consequence of this finding that the ordering is not exactly the same? Wouldn't we expect it to not be the same? Or is the example you added of the black color filter being the top filter in all networks our best attempt at motivating why someone should care?}
Similar differences exist when observing the learned edge filters: in \net{1} and \net{4} the most active edge filter is horizontal; in \net{2} and \net{3} it is vertical. The right half of \figref{highest_lowest_conv1} depicts the least active filters. The same lack of alignment arises, but here the activation values are more tightly packed, so the exact ordering is less meaningful.
\figref{highest_lowest_conv2} shows the even more widely varying activations from \layer{conv2}.

\begin{figure}[htbp]
\begin{center}
 \includegraphics[width=1\linewidth]{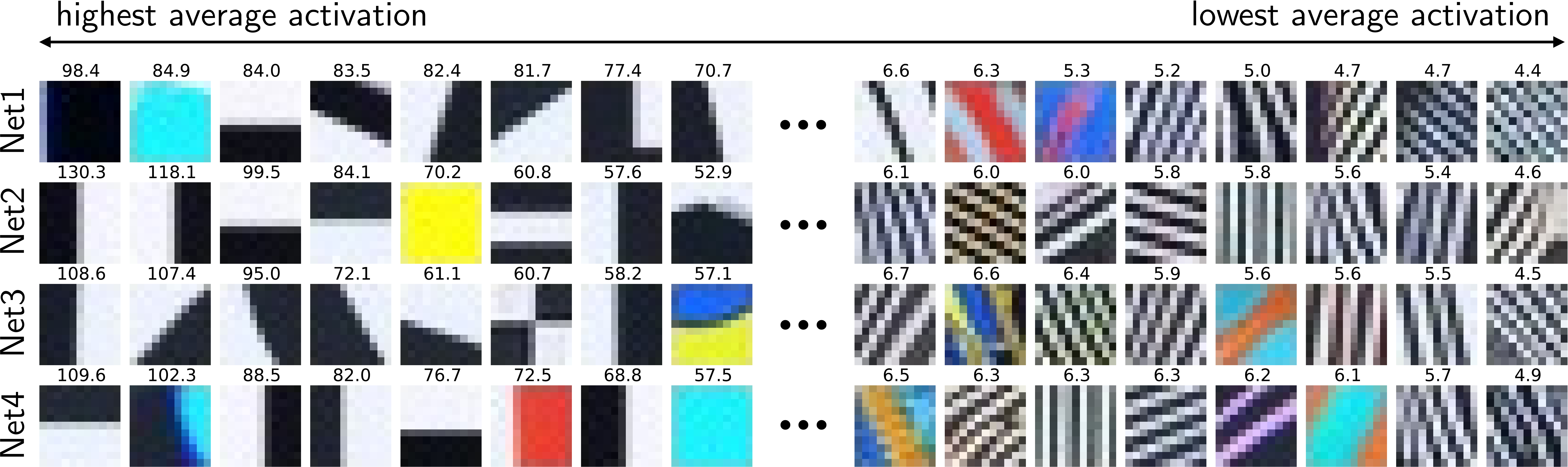}
 \caption{The most active (left) to least active (right) \layer{conv1} filters from \net{1} -- \net{4}, with average activation values printed above each filter.
   The most active filters generally respond to low spatial frequencies, and the least active filtered to high spatial frequencies, but the lack of alignment is interesting (see text).}
 \figlabel{highest_lowest_conv1}
\end{center}
\vskip -0.2in
\end{figure}

\begin{figure}[htbp]
\vskip 0.2in
\begin{center}
  \includegraphics[width=1\linewidth]{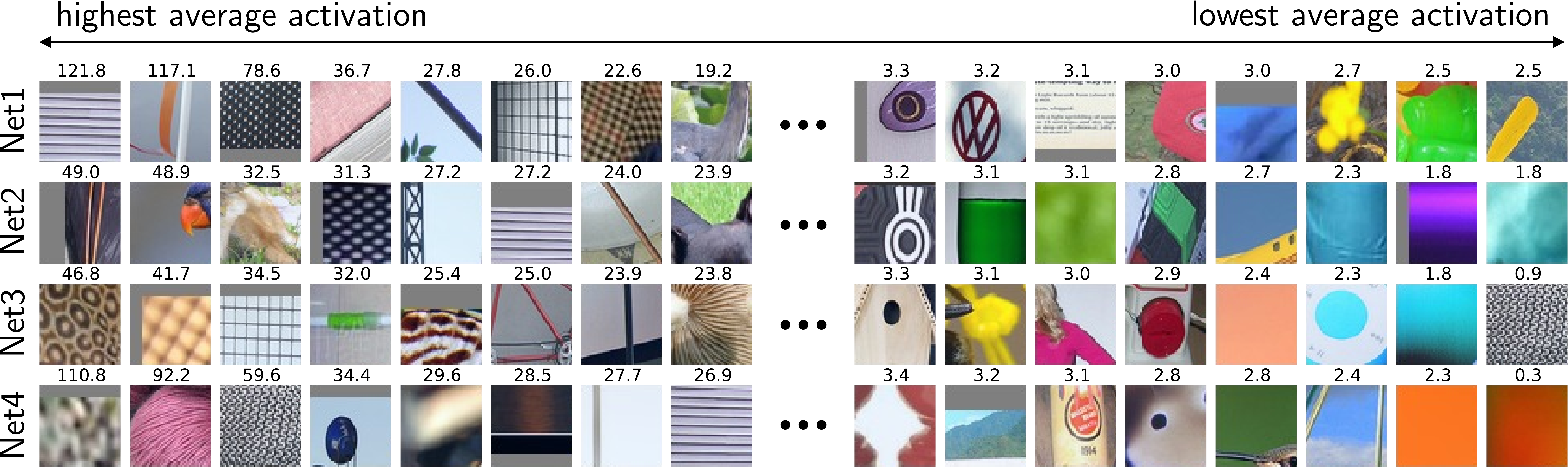}
  \caption{Most active (left) to least active (right) \layer{conv2} filters as in \figref{highest_lowest_conv1}. Compared to the \layer{conv1} filters, here the separation and misalignment between the top filters is even larger. For example, the top unit responding to horizontal lines in \net{1} has average activation of 121.8, whereas similar units in \net{2} and \net{4} average 27.2 and 26.9, respectively. The unit does not appear in the top eight units of \net{3} at all. The least active units seem to respond to rare specific concepts.}
  \figlabel{highest_lowest_conv2}
\end{center}
\end{figure}

\end{document}

%% file: authors.tex
\author{
Yixuan Li\textsuperscript{\textrm{1}}\thanks{Authors contributed equally.},~
Jason Yosinski\textsuperscript{\textrm{1}}\footnotemark[1],~
Jeff Clune\textsuperscript{\textrm{2}},~
Hod Lipson\textsuperscript{\textrm{3}},~ \&
John Hopcroft\textsuperscript{\textrm{1}} \\ 
\textsuperscript{1}Cornell University \\
\textsuperscript{2}University of Wyoming \\
\textsuperscript{3}Columbia University \\
\texttt{\{yli,yosinski,jeh\}@cs.cornell.edu} \\
\texttt{jeffclune@uwyo.edu, hod.lipson@columbia.edu}
}